\documentclass[sn-mathphys]{sn-jnl}

\newtheorem{theorem}{Theorem}
\makeatletter
\newcommand{\multiline}[1]{%
  \begin{tabularx}{\dimexpr\linewidth-\ALG@thistlm}[t]{@{}X@{}}
    #1
  \end{tabularx}
}
\makeatother

\usepackage{subcaption,graphicx}
\usepackage{placeins}
\usepackage{bm}
\usepackage{tabularx}
\usepackage{enumitem}
\usepackage[export]{adjustbox}
\usepackage{natbib}
\algnewcommand{\LineComment}[1]{\State \(\triangleright\) #1}
 
\newcommand{\n}{{\hat{n}}} 
 
\newcommand{\R}{{\mathbb{R}}} 
 
\newcommand{\N}[2]{{\mathcal{N}\left(#1, #2\right)}}

\newcommand{\x}{X}
\DeclareMathOperator*{\argmax}{arg\,max}

\begin{document}
\title[Likelihood-ratio-based confidence intervals]{Likelihood-ratio-based confidence intervals for neural networks}

\author*[1]{\fnm{Laurens} \sur{Sluijterman}}\email{l.sluijterman@math.ru.nl}

\author[1]{\fnm{Eric} \sur{Cator}}\email{e.cator@science.ru.nl}
%
\author[2]{\fnm{Tom} \sur{Heskes}}\email{tom.heskes@ru.nl}
%
\affil[1]{\orgdiv{Department of Mathematics}, \orgname{Radboud University}, \orgaddress{\city{Nijmegen}}}
\affil[2]{\orgdiv{Institute for Computing and Information Sciences}, \orgname{Radboud University}, \orgaddress{\city{Nijmegen}}}

%
%


\abstract{This paper introduces a first implementation of a novel likelihood-ratio-based approach for constructing confidence intervals for neural networks. Our method, called DeepLR, offers several qualitative advantages: most notably, the ability to construct asymmetric intervals that expand in regions with a limited amount of data, and the inherent incorporation of factors such as the amount of training time, network architecture, and regularization techniques. While acknowledging that the current implementation of the method is prohibitively expensive for many deep-learning applications, the high cost may already be justified in specific fields like medical predictions or astrophysics, where a reliable uncertainty estimate for a single prediction is essential. This work highlights the significant potential of a likelihood-ratio-based uncertainty estimate and establishes a promising avenue for future research.}


\maketitle

\begin{figure}[h]
\centering
\begin{subfigure}{0.33\textwidth}
  \centering
  \includegraphics[width=\linewidth]{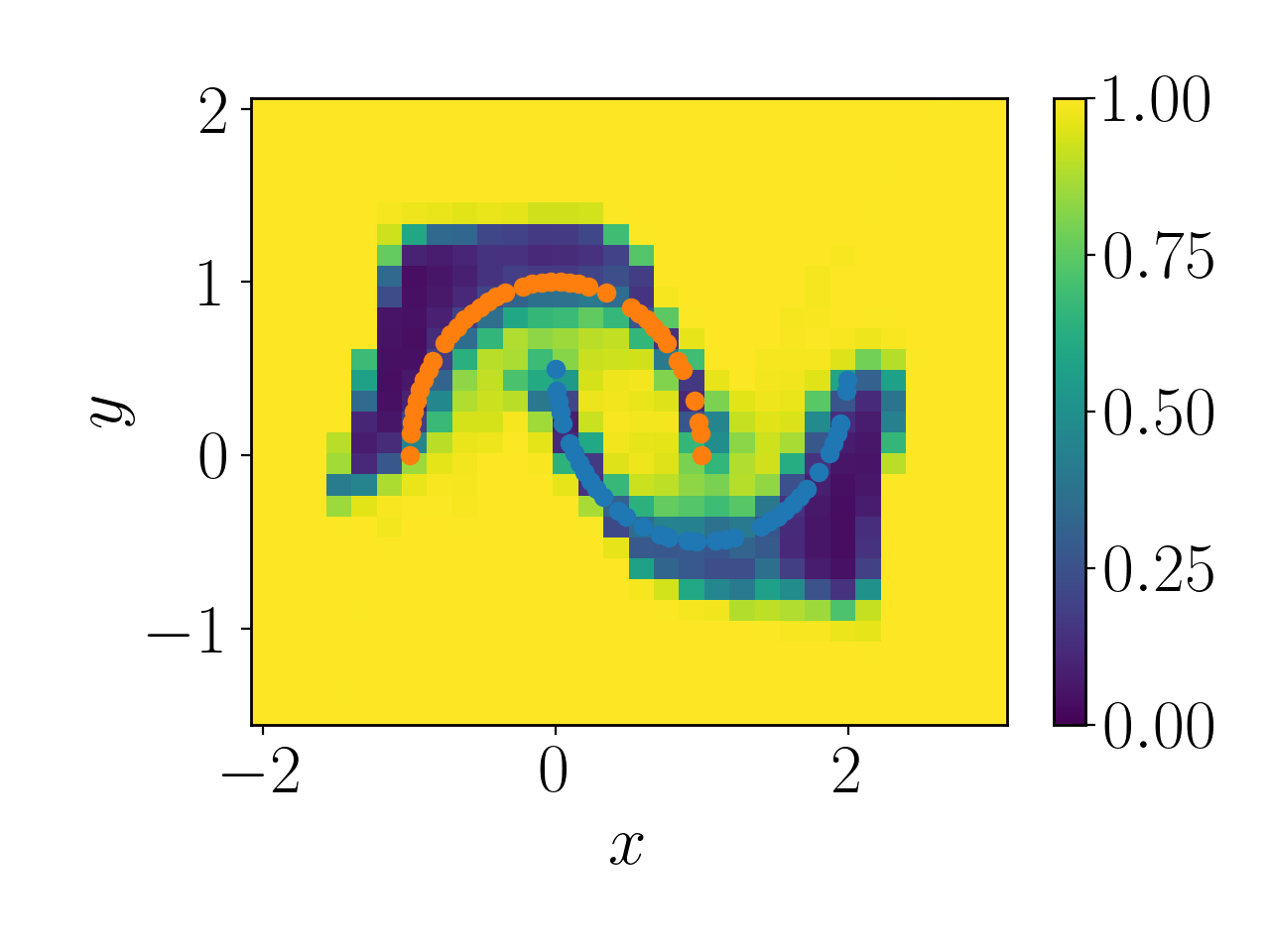}
  \caption{DeepLR}
  \label{fig:1}
\end{subfigure}
\begin{subfigure}{0.33\textwidth}
  \centering
  \includegraphics[width=\linewidth]{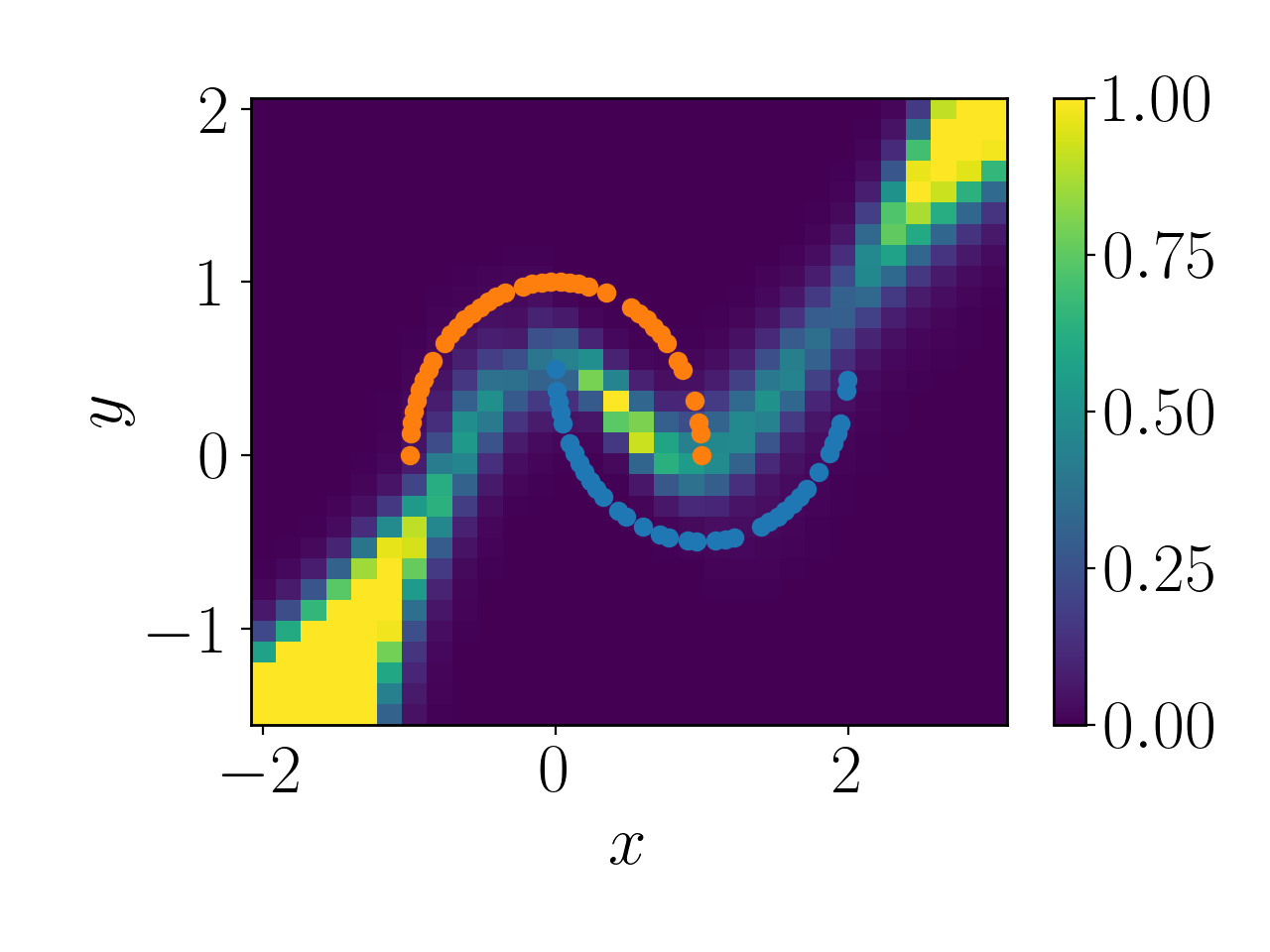}
  \caption{Ensembling}
  \label{fig:1}
\end{subfigure}
\begin{subfigure}{0.325\textwidth}
  \centering
  \includegraphics[width=\linewidth]{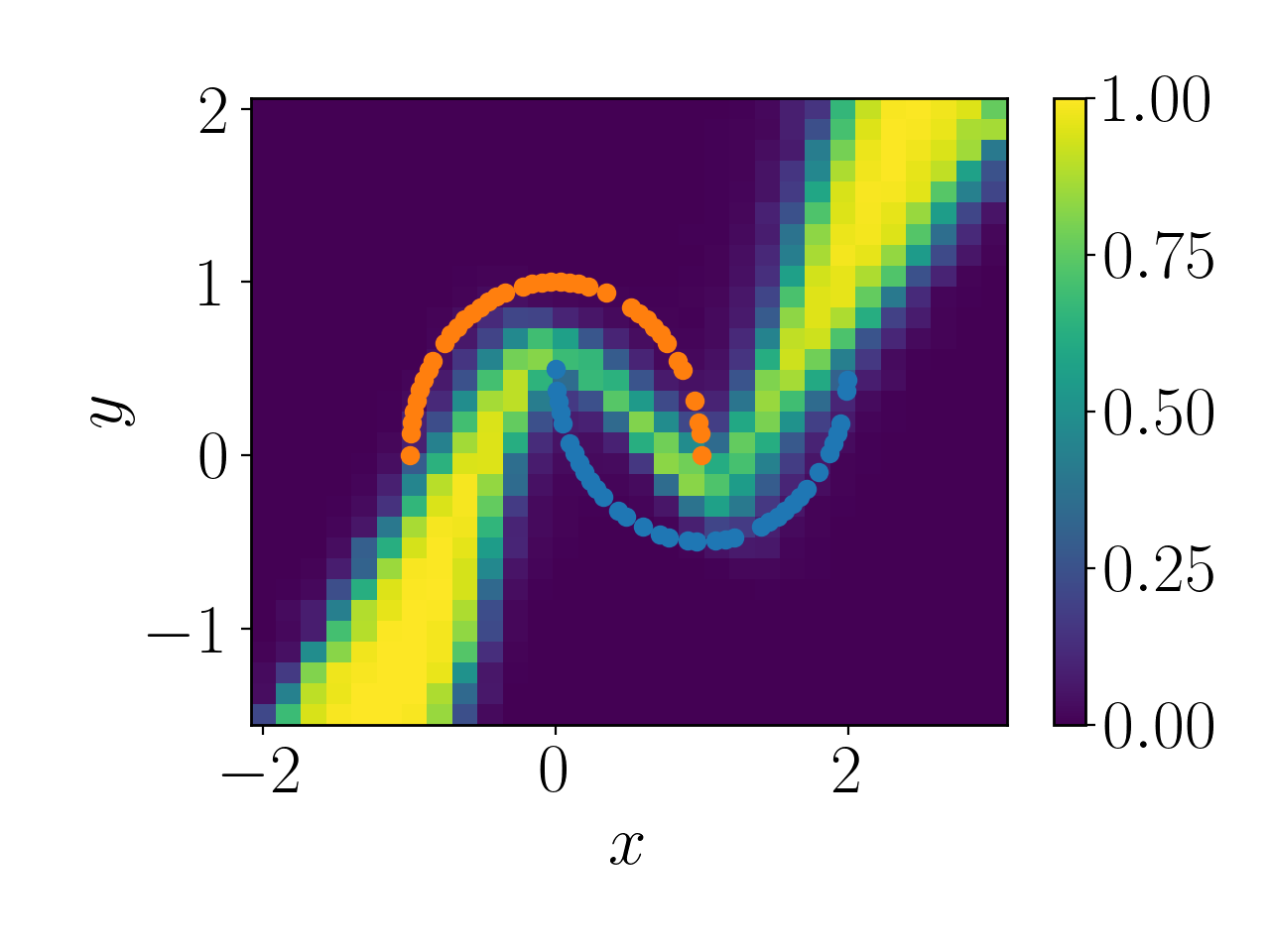}
  \caption{Dropout}
  \label{fig:1}
\end{subfigure}
\caption{A comparison of the confidence intervals of our likelihood-ratio approach (DeepLR), an ensembling approach, and MC-Dropout on the two-moon data set. The colorbar represents the width of 95$\%$ confidence intervals, where yellow indicates greater uncertainty. The orange and the blue circles indicate the location of the data points of the two different classes. Crucially, DeepLR presents high levels of uncertainty in regions far away from the data, generating confidence intervals of [0.00, 1.00], unlike the other methods that display extreme certainty in those regions.}
\label{fig: twomoon}
\end{figure}
\section{Introduction}\label{sec1}
Over the past two decades, neural networks have seen an enormous rise in popularity and are currently being used in almost every area of science and industry. In light of this widespread usage, it has become increasingly clear that trustworthy uncertainty estimates are essential \citep{gal2016uncertainty}.

Many uncertainty estimation methods have been developed using Bayesian techniques \citep{neal2011mcmc, mackay1992practical, gal2016uncertainty}, ensembling techniques \citep{heskes1997practical, lakshminarayanan2017simple}, or applications of frequentist techniques such as the delta method \citep{kallus2022Implicit}. 

Many of the resulting confidence intervals (for the frequentist methods) and credible regions (for the Bayesian methods) have two common issues. Firstly, most methods result in symmetric intervals around the prediction which can be overly restrictive and can lead to very low coverage in biased regions \citep{sluijterman2022confident}. Secondly, most methods rely heavily on asymptotic theorems (such as the central limit theorem or the Bernstein-von-Mises theorem) and can therefore only be trusted in the asymptotic regime where we have many more data points than model parameters, the exact opposite scenario of where we typically find ourselves within machine learning.

\paragraph{Contribution} 
\noindent In this paper, we demonstrate how the likelihood-ratio test can be leveraged to combat the two previously mentioned issues. We provide a first implementation of a likelihood-ratio-based approach, called DeepLR, that has the ability to produce asymmetric intervals that are more appropriately justified in the scenarios where we have more parameters than data points. Furthermore, these intervals exhibit desirable behavior in regions far removed from the data, as evidenced in Figure \ref{fig: twomoon}.

\paragraph{Organisation}
This paper is structured into four sections, with this introduction being the first. Section \ref{method} explains our method in detail and also contains the related work section which is simultaneously used to highlight the advantages and disadvantages of our method. Section \ref{results} presents experimental results that illustrate the desirable properties of a likelihood-ratio-based approach. Finally, Section \ref{discussionconclusion} summarizes and discusses the results and outlines possible directions for future work.

\FloatBarrier

\section{DeepLR: Deep Likelihood-Ratio-based confidence intervals} \label{method}
In this section, we present our method, named DeepLR, for constructing confidence intervals for neural networks using the likelihood-ratio test. We first formalize the problem that we are considering in Subsection \ref{Problem formulation}. Subsection \ref{llrtest} explains the general idea behind constructing a confidence interval via the likelihood-ratio test. Subsequently, in Subsection \ref{high level idea}, we outline the high level idea for translating this general procedure to neural networks. The details regarding the distribution and the calculation of the test statistic are provided in Subsections \ref{distribution of test statistic} and \ref{obtaining test statistic}. Finally, in Subsection \ref{related work}, we compare our method to related work while simultaneously highlighting its strengths and limitations.

\subsection{Problem formulation} \label{Problem formulation}
We consider a data set $\mathcal{D} = \{(X_{1}, Y_{1}), \ldots, (\x_{n}, Y_{n}) \}$, consisting of $n$ independent observations of the random variable pair $(X, Y)$. We consider networks that provide an estimate for the conditional density of $Y \mid X$. This is achieved by assuming a distribution and having the network output the parameter(s) of that distribution. 

Three well-known types of networks that fall in this class are: 
(1) A regression setting where the network outputs a mean estimate and is trained using a mean-squared error loss. This is equivalent to assuming a normal distribution with homoscedastic noise. 
(2) Alternatively, the network could output both a mean and a variance estimate and be optimized by minimizing the negative loglikelihood assuming a normal distribution. 
(3) In a classification setting, the network could output logits that are transformed to class probabilities while assuming a categorical distribution.

The network is parametrized by $\theta \in \mathbb{R}^{p}$, where  $p$ is typically much larger than $n$.  With $\Theta$, we denote the set containing all the $\theta$ that are reachable for a network with a specific training process. This includes choices such as training time, batch size, optimizer, and regularization techniques. With $p_{\theta}$, we denote the predicted conditional density. Additionally, we assume that the true conditional density is given by $p_{\theta_{0}}$ for some $\theta_{0}$ in $\Theta$. In other words, we assume that our model is well specified. 

The objective of our method is to construct a confidence interval for one of the output nodes of the network for a specific input of interest $X_{0}$. We denote this output of interest with $f_{\theta_{0}}(X_{0})$ for the remainder of the paper. In the context of a regression setting, this output of interest is the true regression function value at $X_{0}$ and in a classification setting it is the true class probability for input $X_{0}$.

 We define a $(1-\alpha) \cdot 100\%$ confidence interval for $f_{\theta_{0}}(X_{0})$ as an interval, $\text{CI}(f_{\theta_{0}}(X_{0}))$, which is random since it depends on the random realization of the data, such that the probability (taken with respect to the random data set) that $\text{CI}(f_{\theta_{0}}(X_{0}))$ contains the true value $f_{\theta_{0}}(X_{0})$ is $(1-\alpha)\cdot 100\%$.

\subsection{Confidence interval based on the likelihood ratio} \label{llrtest}
We explain the general idea behind constructing a confidence interval with the likelihood-ratio by working through a well-known example. We consider $n$ observations $Y_{i}$ that are assumed to be normally distributed with unknown mean $\mu$ and unknown variance $\sigma^{2}$. Our goal is to create a confidence interval for $\mu$ by using the likelihood-ratio test. 

The duality between a confidence interval and hypothesis testing states that we can create a $(1-\alpha)\cdot 100\%$ confidence interval for $\mu$ by including all the values $c$ for which the hypothesis $\mu=c$ cannot be rejected at a $(1-\alpha)\cdot 100\%$ confidence level. We must therefore test for what values $c$ we can accept the hypothesis $\mu=c$.

The general approach to test a hypothesis is to create a test statistic of which we know the distribution under the null hypothesis and to reject this hypothesis if the probability of finding the observed test statistic or an extremer value is smaller than $\alpha$. 

As our test statistic, we take two times the log of the likelihood ratio:
\[
T(c) := 2\left(\sup_{\Theta}\left(\sum_{i=1}^{n}\log(L(Y_{i};\theta))\right) - \sup_{\Theta_{0}}\left(\sum_{i=1}^{n}\log(L(Y_{i}; \theta))\right)\right),
\]
where $L$ denotes the likelihood function $\theta \mapsto L(Y_{i} ;\theta)$, $\Theta$ is the full parameter space and $\Theta_{0}$ the restricted parameter space. In our example, we have
\[
\Theta = \{(\mu, \sigma^{2}) \mid \mu \in \mathbb{R}, \sigma^{2} \in \mathbb{R}_{>0}\},
\]
and
\[
\Theta_{0} = \{(c, \sigma^{2}) \mid \sigma^{2} \in \mathbb{R}_{>0}\}.
\]

\citet{wilks1938LargeSample} proved that $T(c)$ weakly converges to a $\chi^{2}(1)$ distribution under the null hypothesis that $\mu=c$. We therefore reject if $T(c) > \chi^{2}_{1-\alpha}(1)$ and our confidence interval for $\mu$ becomes the set $\{c \mid T(c) \leq \chi^{2}_{1-\alpha}(1)\}$, where $\chi^{2}_{1-\alpha}(1)$ is the $(1-\alpha)$-quantile of a $\chi^{2}(1)$ distribution.

In our example, this results in the well-known interval 

\[
\bar{Y} \pm z_{1-\alpha/2}\sqrt{ \frac{1}{n} \frac{1}{n-1}\sum_{i=1}^{n}(Y_{i} - \bar{Y})^{2}},
\]
where $z_{1-\alpha/2}$ is the $(1-\alpha/2)$-quantile of a standard-normal distribution.
\subsection{High-level idea of DeepLR} \label{high level idea}
Our goal is to apply the likelihood-ratio testing procedure outlined in the previous subsection to construct a confidence interval for $f_{\theta_{0}}(X_{0})$, the value of one of the output nodes given input $X_{0}$. We create this confidence interval by including all the values $c$ for which the hypothesis $f_{\theta_{0}}(X_{0})=c$ cannot be rejected. The testing of the hypothesis is done with the likelihood-ratio test. Specifically, we use two times the log likelihood ratio as our test statistic:
\begin{align}
T(c) &:= 2\bigg(\sup_{\Theta}\big(\sum_{i=1}^{n}\log(L(X_{i}, Y_{i};\theta))\big) - \sup_{\Theta_{0}(c)}\big(\sum_{i=1}^{n}\log(L(X_{i}, Y_{i};\theta))\big)\bigg) \nonumber \\
&= 2\bigg(\sup_{\Theta}\big(\sum_{i=1}^{n}\log(p_{\theta}(Y_{i} \mid X_{i}))\big) - \sup_{\Theta_{0}(c)}\big(\sum_{i=1}^{n}\log(p_{\theta}(Y_{i} \mid X_{i}))\big)\bigg),
\label{eq: teststatistic} 
\end{align}
and we construct a confidence interval for $f_{\theta_{0}}(X_{0})$ by including all values $c$ for which the test statistic is not larger than $\chi^{2}_{1-\alpha}(1)$:
\begin{equation}
\text{CI}(f_{\theta_{0}}(X_{0})) = \{c \mid T(c) \leq \chi^{2}_{1-\alpha}(1)\}.
\label{eq: CI}
\end{equation}

Here, we consider $\Theta \subset \mathbb{R}^{p}$ to be the set containing all reachable parameters, and $\Theta_{0}(c) =\{\theta \in \Theta \mid f_{\theta_{0}}(X_{0}) = c\}$. The set $\Theta$ is explicitly not equal to all parameter combinations. Due to explicit (e.g., early stopping) and implicit regularization (e.g., lazy training \citep{chizat2019lazy}) not all parameter combinations can be reached. The set $\Theta$ should therefore be seen as the set containing the parameters of all neural networks that can be found given the optimizer, training time, regularization techniques, and network architecture.

Intuitively, this approach answers the question: \textit{What values could the network have reached at location $X_{0}$ while still explaining the data well?} This is a sensible question for a highly flexible and typically overparameterized machine-learning approach. After training, the model ends up with a certain prediction at location $X_{0}$. However, since the model is typically very complex, it is likely that the model could just as well have made other predictions at that location while still explaining the data well. Therefore, all those other function values should also be considered as possibilities. Inherently, all modeling choices are taken into account by asking this question. A more flexible model, for instance, is likely able to reach more values without affecting the likelihood of the training data, leading to a larger confidence interval.

The construction of the confidence interval in Equation \eqref{eq: CI} assumes that the test statistic, $T(c)$, has a $\chi^{2}(1)$ distribution. We discuss this assumption in the following subsection. The subsection thereafter describes how to calculate the test statistic.

\subsection{Distribution of the test statistic} \label{distribution of test statistic}
In the classical setting, \citet{wilks1938LargeSample} proved that the likelihood-ratio test statistic asymptotically has a $\chi^{2}(1)$ distribution when the submodel has one degree of freedom less than the full model. We are, however, not in this classical regime. We have many more parameters than data points and therefore need a similar result for this setting.  

It has been shown that the likelihood-ratio test statistic converges to a $\chi^{2}$ distribution for a wide range of settings, which is referred to as the Wilks-phenomenon by later authors \citep{fan2001Generalized, boucheron2011Highdimensional}. For a semi-parametric model, which more closely resembles our situation, it has been shown that the test statistic also converges in distribution to a $\chi^{2}$ distribution under appropriate regularity conditions \citep{murphy1997Semiparametric}. 

We prove a similar result for our setting in the appendix. The theorem states that, under suitable assumptions, our test statistic has a $\chi^{2}(1)$ distribution. Intuitively, this results from the fact that we added a single constraint, namely that $f_{\theta}(X_{0})=c$. We emphasize that even if the test statistic does not exactly follow a $\chi^{2}(1)$ distribution, the qualitative characteristics of the confidence intervals will remain evident, albeit with inaccurate coverage levels. 

\subsection{Calculating the test statistic} \label{obtaining test statistic}
Calculating the two terms in equation \eqref{eq: teststatistic} presents certain challenges. The first term, $\sup_{\Theta}\big(\sum_{i=1}^{n}\log(p_{\theta}(Y_{i} \mid X_{i}))\big)$, is relatively straightforward. We train a network that maximizes the likelihood, which gives the conditional densities $p_{\hat{\theta}}(Y_{i} \mid X_{i})$. 

The second term is substantially more complex. Ideally, we would optimize over the set $\Theta_{0}(c) = \{\theta \in \Theta \mid f_{\theta}(X_{0})=c\}$. This is problematic for two reasons. Firstly, it is unclear how we can add this constraint to the network and secondly, this would necessitate training our network for every distinct value $c$ that we wish to test. Even when employing an efficient bisection algorithm, this could easily result in needing to train upwards of 10 additional networks. 

\begin{figure}[h]
\centering
\begin{subfigure}{0.49\textwidth}
  \centering
  \caption*{Step 1: Train a network on the data.}
  \vskip 0.45in
  \includegraphics[width=\linewidth]{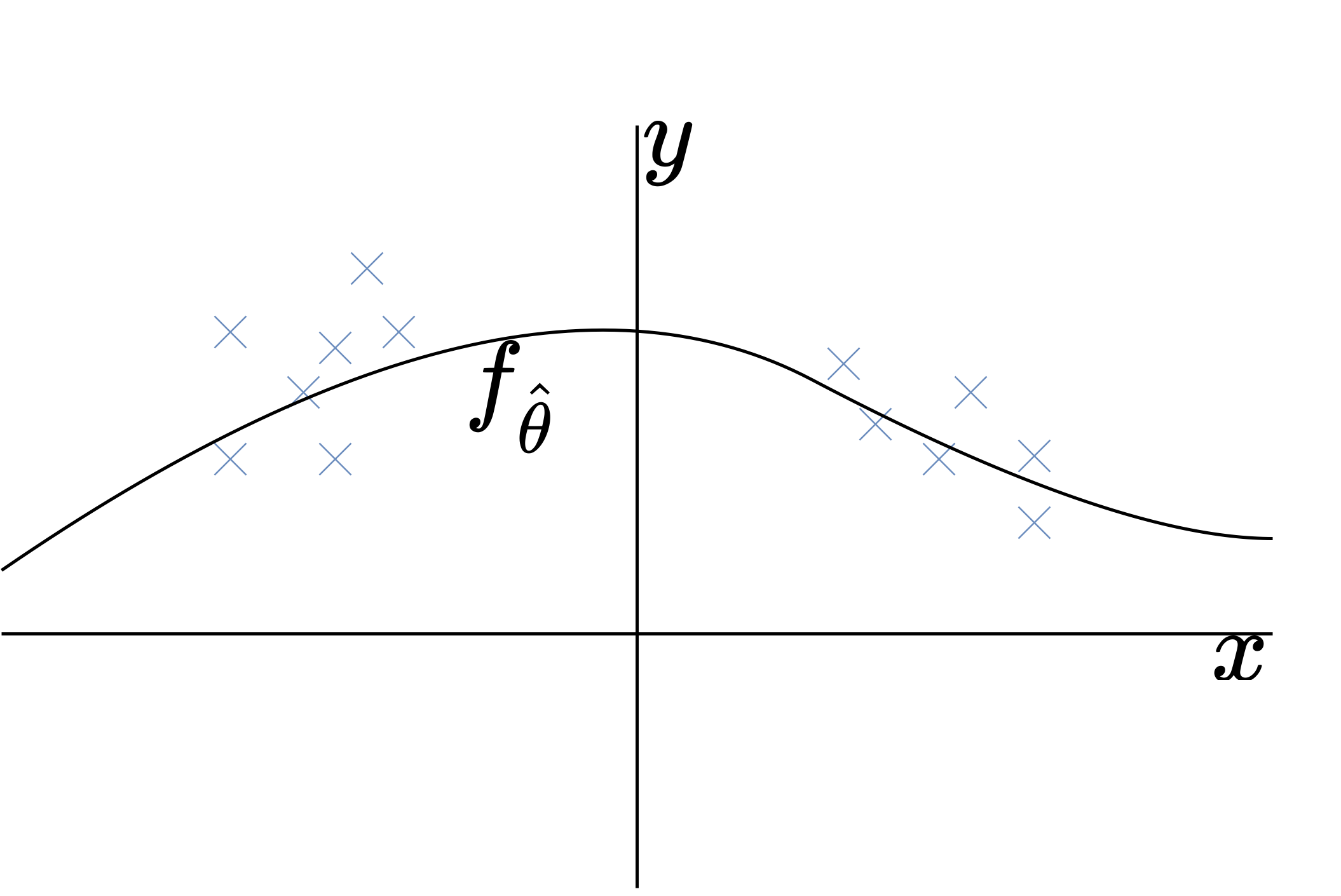}
\end{subfigure}
\hfill
\begin{subfigure}{0.49\textwidth}
  \centering
  \caption*{Step 2: Copy the resulting network and train it on the objective to have the same predictions at the training locations and a larger prediction at $X_{0}$.}
  \includegraphics[width=\linewidth]{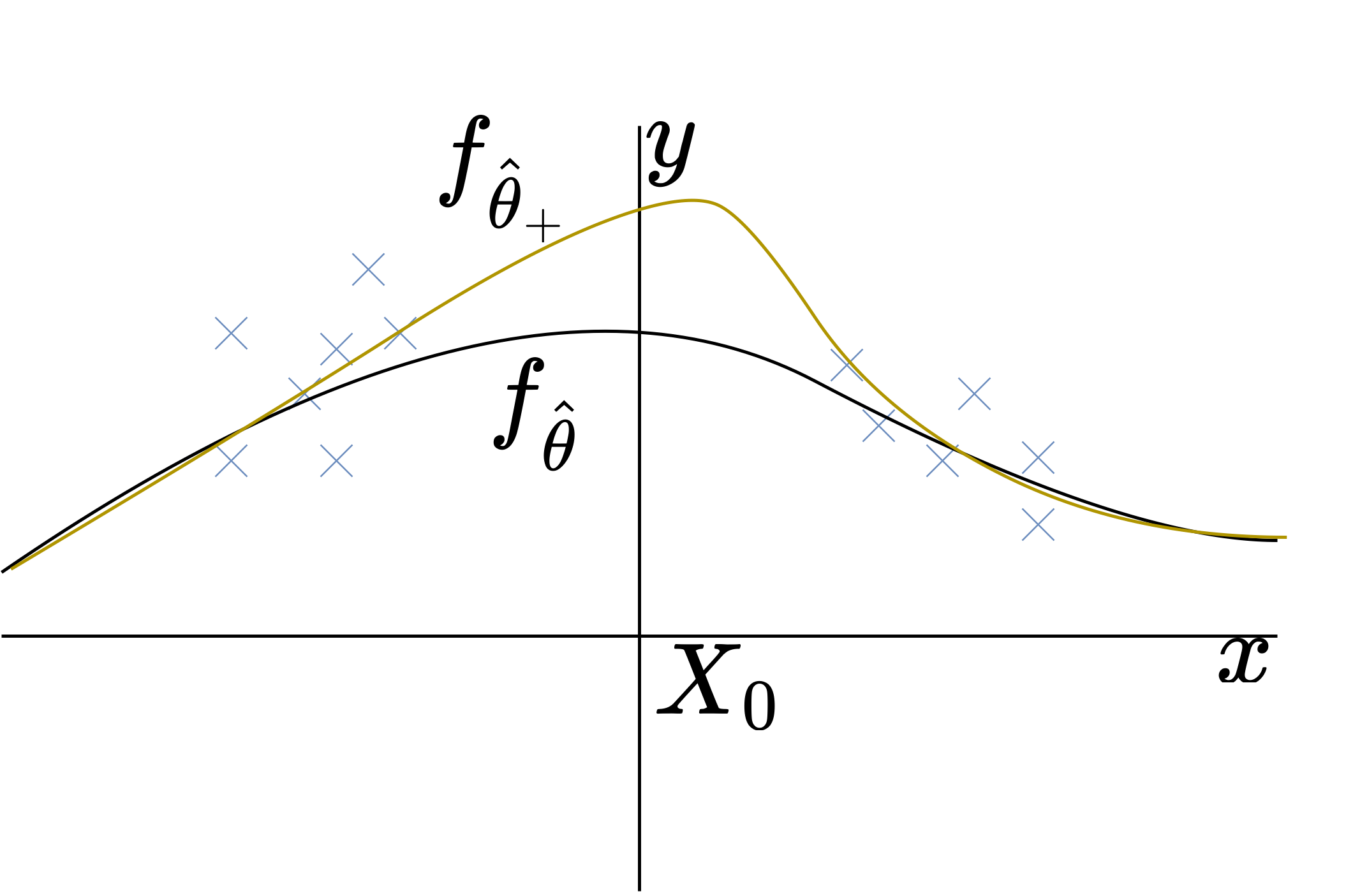}
\end{subfigure}

\vspace{0.5cm}

\begin{subfigure}{0.49\textwidth}
  \centering
  \caption*{Step 3: Choose $\lambda$ such that \\ $(1-\lambda) f_{\hat{\theta}}(X_0) + \lambda f_{\hat{\theta}_{+}}(X_{0})=c$ and let $p_c$ be the density resulting from the same linear combination of the predicted distributional parameters.}
  \includegraphics[width=\linewidth]{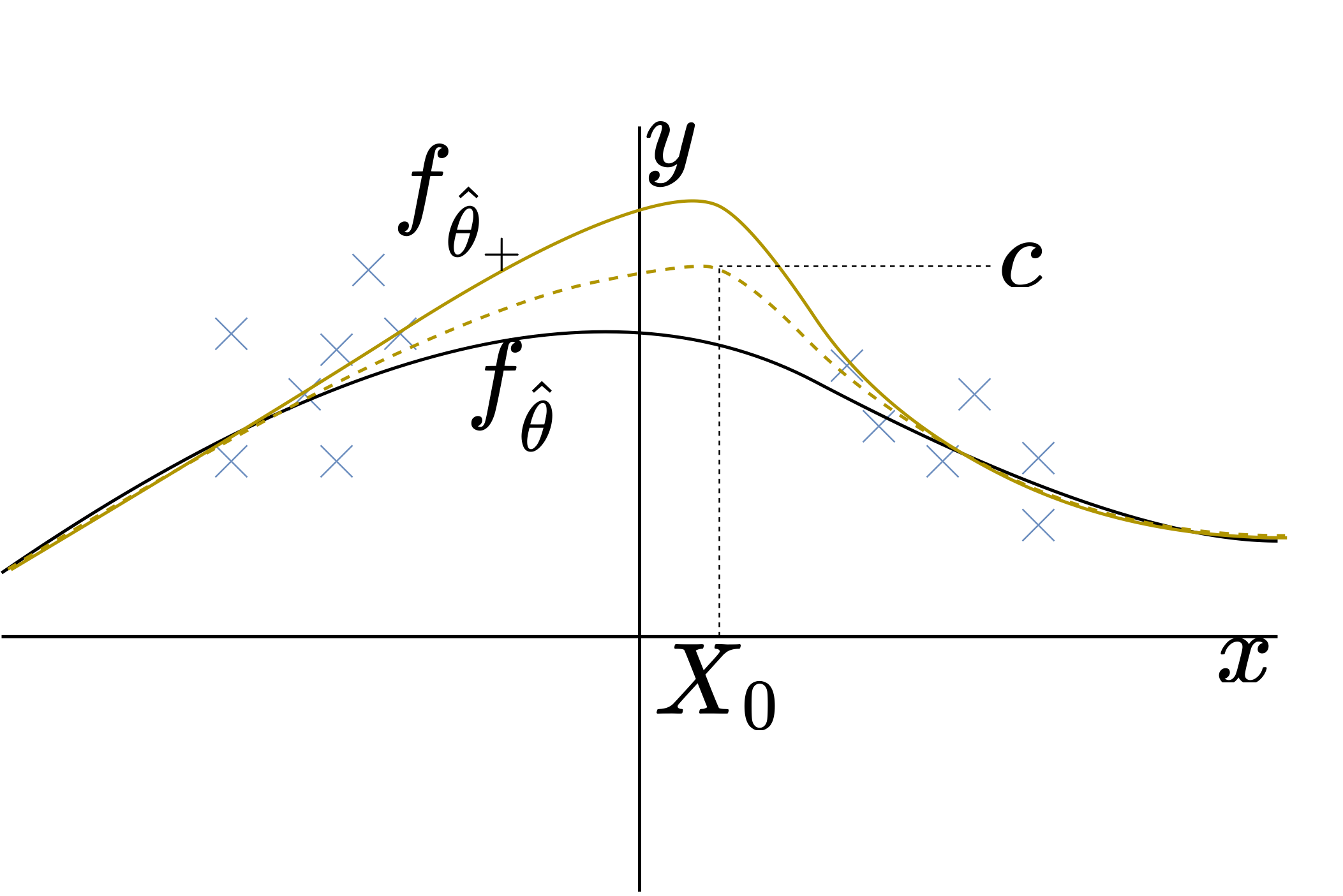}
\end{subfigure}
\hfill
\begin{subfigure}{0.49\textwidth}
  \centering
  \vskip -0.35in
  \caption*{Step 4: Repeat step 3 for multiple values of $c$ and define the confidence interval for $\text{CI}(f_{\theta_{0}}(X_{0}))$ as all $c$ for which:}
  \vskip 0.3in
  \begin{minipage}{1\textwidth}
  	\begin{align*}
  		& 2\bigg(\sum_{i=1}^{n}\log(p_{\hat{\theta}}(Y_{i} |X_{i})) \\
  		& \quad \quad \quad - \sum_{i=1}^{n}\log(p_{c}(Y_{i} \mid X_{i}))\bigg)\\
  		&\quad \quad \quad  \leq \chi^{2}_{1-\alpha}(1).
  	\end{align*}
  \end{minipage}
\end{subfigure}
\vskip 0.1in
\caption{Illustration of the steps of our method for the positive direction in a regression setting. Steps 2, 3, and 4 are also carried out in the negative direction.}
\label{fig: allsteps}
\end{figure}

We address this problem as follows. We first create a network that is perturbed in the direction of a relatively large value ($c_{\text{max}}$) at $X_{0}$ and a network that is perturbed in the direction a relatively small value $(c_{\text{min}})$ at $X_{0}$ while maximizing the likelihood of the data. We denote the resulting network parameters with $\hat{\theta}_{\pm}$:
\[
\hat{\theta}_{+} = \argmax_{\underset{f_{\theta}(X_{0})\approx c_{\text{max}}}{\theta \in \Theta,}} L(\mathcal{D};\theta), \quad \text{and} \quad \hat{\theta}_{-} = \argmax_{\underset{f_{\theta}(X_{0})\approx c_{\text{min}}}{\theta \in \Theta,}} L(\mathcal{D};\theta).
\]
Subsequently, the network that maximizes the likelihood under the constraint $f_{\theta}(X_{0})=c$ is approximated using a linear combination. Specifically, suppose we want the network that maximizes the likelihood of the training data while passing through $c$ at $X_{0}$. In the case that $c>f_{\hat{\theta}}(X_{0})$, we approximate this network by taking a linear combination of the outputs such that
\[
(1-\lambda) f_{\hat{\theta}}(X_{0})+ \lambda f_{\hat{\theta}_{+}}(X_{0}) = c,
\]
and we define $p_{c}$ as the density that we get by using the same linear combinations for the distributional parameters that are predicted by the networks parametrized by $\hat{\theta}$ and $\hat{\theta}_{+}$. We then approximate the second term in equation \eqref{eq: teststatistic} as follows:
\begin{equation} \label{approximation}
 \sup_{\Theta_{0}(c)}\big(\sum_{i=1}^{n}\log(p_{\theta}(Y_{i} \mid X_{i}))\big) \approx \sum_{i=1}^{n} \log(p_{c}(Y_{i} \mid X_{i})).
\end{equation}
This procedure is visualized in Figure \ref{fig: allsteps} for a regression setting. Details on the second step, finding the perturbed networks, are provided below both for a regression and binary-classification setting.

\paragraph{Regression}
Suppose we completed step 1 and we have a network that maximizes the data's likelihood. For step 2, our objective is to adjust this network such that it goes through a relatively large value at $X_{0}$ while continuing to maximize the data's likelihood. Moreover, we aim to achieve this in as stable a manner as possible, given that minor differences can significantly influence the test statistic, particularly for large data sets. 

We accomplish this by copying the original network and training it on the objective to maintain the original predictions --  as those maximized the likelihood -- while predicting $f_{\hat{\theta}}(X_{0}) + 1$ for input $X_{0}$. For the perturbation in the negative direction we use $-1$. These values $\pm 1$ are chosen assuming that the targets are normalized prior to training.

We obtain this training objective by using modified training set, $\tilde{\mathcal{D}}$, that is constructed by replacing the targets $Y_{i}$ in the original training set with the predictions $f_{\hat{\theta}}(X_{i})$ of the original network and adding the data point $(X_{0}, f_{\hat{\theta}}(X_{0})  +1).$

The resulting problem is very imbalanced. We want the network to change the prediction at location $X_{0}$, which is only present in the data once. This makes the training very unstable since, especially when training for a small number of epochs, it is very influential which specific batch contains the new point. To remedy this, we use a combination of upsampling and downweighting of the new data point $(X_{0}, f_{\hat{\theta}}(X_{0})  +1).$ Merely upsampling the new data point -- i.e., adding it multiple times -- is undesirable as this can introduce significant biases \citep{vandengoorbergh2022Harm}. Hence, we add many copies of $(X_{0}, f_{\hat{\theta}}(X_{0}) + 1)$ but we reweigh the loss contributions of these added data points such that they have a total contribution to the loss that is equivalent to that of a single data point. 

We propose to add $2n/\text{batch size}$ extra data points such that each batch is expected to have 2 new data points. The same training procedure is used as for the original network. We found slightly larger or smaller number of added data points to perform very similarly. This setting worked for a wide variety of data sets and architectures.

\paragraph{Binary classification}
Consider a data set where the targets are either 1 or 0. Our network outputs logits, denoted with $f_{\hat{\theta}}(\x)$, that are transformed to probabilities via a sigmoid function. The procedure is nearly identical for this binary classification setting: We create a positively perturbed network, parametrized by $\hat{\theta}_{+}$, and a negatively perturbed network, parametrized by $\hat{\theta}_{-}$.

The only difference is in the construction of the augmented data sets. We again replace the targets $Y_{i}$ by the predictions of the original network, $f_{\hat{\theta}}(\x_{i})$, but now we add multiple copies of the data point $(X_{0}, 1)$ for the positive direction, and multiple copies of the data point $(X_{0}, 0)$ for the negative direction. 

The entire method is summarized in Algorithm \ref{alg: entiremethod}. In summary, we want to test what values the network can reach while still explaining the data well. We do this by perturbing the network in a positive direction and a negative direction and subsequently testing which linear combinations would still explain the data reasonably well, i.e., linear combinations with a test statistic smaller than $\chi^{2}_{1-\alpha}(1)$.

\begin{algorithm}[h]
\caption{Pseudocode for the construction of $\text{CI}(f_{\theta_{0}}(X_{0}))$ using DeepLR}\label{algo1}
\begin{algorithmic}[1]
\Require $\mathcal{D} = \{(X_{1}, Y_{1}), \ldots, (X_{n}, Y_{n}) \}$, $\hat{\theta}, X_{0}$, $\alpha$;
\State $n_{\text{extra}} = 2 n / (\text{batch size})$; 
\State For binary classification $c_{\text{max}}=1$ and $c_{\text{min}}=0$, and
\Statex for regression $c_{\text{max}} = f_{\hat{\theta}}(X_{0}) + \delta$ and $c_{\text{min}} = f_{\hat{\theta}}(X_{0}) - \delta$;
\State $\tilde{\mathcal{D}}_{+} = \{(\x_{1}, f_{\hat{\theta}}(\x_{1})), \ldots, (\x_{n}, f_{\hat{\theta}}(\x_{n})), \overbrace{(X_{0}, c_{\text{max}}), \ldots, (X_{0}, c_{\text{max}})}^{n_{\text{extra}} \text{ times}}\}$;
\State $\tilde{\mathcal{D}}_{-} = \{(\x_{1}, f_{\hat{\theta}}(\x_{1})), \ldots, (\x_{n}, f_{\hat{\theta}}(\x_{n})), \overbrace{(X_{0}, c_{\text{min}}), \ldots, (X_{0}, c_{\text{min}}}^{n_{\text{extra}} \text{ times}}\}$;
\State Make two copies of network parametrized by $\hat{\theta}$ and train them on $\tilde{\mathcal{D}}_{+}$ and $\tilde{\mathcal{D}}_{-}$ using the original training procedure. During the training, the loss contribution of the added data points is divided by $n_{\text{extra}}$. Denote the parameters of the resulting networks with $\hat{\theta}_{+}$ and $\hat{\theta}_{-}$;
\For{$c \in \mathbb{R}$}  \Comment{Since testing all possible $c$ is impossible, we propose to \Statex \ \ \ \ \ \ \ \ \ \ \ \ \ \ \ \ \ \ \ \ \ \ \ \ \ use some variation of a bisection search algorithm.}
\If{$c > f_{\hat{\theta}}(X_{0})$}
	\State \multiline{%
	Pick $\lambda$ such that $(1-\lambda) f_{\hat{\theta}}(X_{0}) + \lambda f_{\hat{\theta}_{+}}(X_{0})=c$ with corresponding density $p_{c}$; $\triangleright$ The density $p_{c}$ is obtained by taking the same linear \Statex \ \ \ \ \ \ \ \ \ \ \ \ \ \ \ \ \ combination of the predicted distribution parameters \Statex \ \ \ \ \ \ \ \ \ \ \ \ \ \ \ \ \ \ outputted by the networks parametrized by $\hat{\theta}$ and $\hat{\theta}_{+}$.}
	\EndIf
	\If{$c < f_{\hat{\theta}}(X_{0})$}
	\State \multiline{%
	Pick $\lambda$ such that $(1-\lambda) f_{\hat{\theta}}(X_{0}) + \lambda f_{\hat{\theta}_{-}}(X_{0})=c$ with corresponding density $p_{c}$;}
	\EndIf
\If{$2\bigg(\sum_{i=1}^{n}\log(p_{\hat{\theta}}(Y_{i} \mid X_{i})) - \sum_{i=1}^{n}\log(p_{c}(Y_{i} \mid X_{i}))\bigg) \leq \chi^{2}_{1-\alpha}(1)$}
\State Include $c$ in $\text{CI}(f_{\theta_{0}}(\x_{0}));$
\EndIf
\EndFor
\State \Return $\text{CI}(f_{\theta_{0}}(X_{0}))$
\end{algorithmic}
\label{alg: entiremethod}
\end{algorithm}

\subsection{Related work} \label{related work}
In this subsection, we place our work in context of existing work while simultaneously highlighting the strengths and limitations of our approach. Our aim is not to give a complete overview of all uncertainty quantification methods, for which we refer to the various reviews and surveys on the subject \citep{abdar2021review, gawlikowski2022Survey, he2023Survey}. Instead, we discuss several broad groups in which most methods can be categorized: Ensembling methods, Bayesian methods, frequentist methods, and distance-aware methods. 

\textit{Ensembling methods} train multiple models and use the variance of the predictions as an estimate for model uncertainty \citep{heskes1997practical, lakshminarayanan2017simple, zhang2017mixup, wenzel2020Hyperparameter, jain2020maximizing, dwaracherla2022ensembles}. While being extremely easy to implement, they can be computationally expensive due to the need to train multiple networks. Moreover, the resulting confidence intervals can behave poorly in regions with a limited amount of data where the predictor is likely biased. Additionally, ensemble members may interpolate in a very similar manner, potentially leading to unreasonably narrow confidence intervals.

\textit{Bayesian approaches} place a prior distribution on the model parameters and aim to simulate from the resulting posterior distribution given the observed data \citep{mackay1992practical, neal2012bayesian, hernandez2015probabilistic}. Since this posterior is generally intractable, it is often approximated, with MC-Dropout being a notable example \citep{gal2016dropout, gal2017concrete}. 

A downside is that these methods can be challenging to train, potentially resulting in a lower accuracy. Our proposed approach does not change the optimization procedure and therefore has no accuracy loss. Another downside is that while asymptotically Bayesian credible sets become confidence sets by the Bernstein-von Mises theorem (see \citet[Chapter 10]{van2000asymptotic})> However, this theorem generally does not apply for a neural network where the dimension of the parameter space typically exceeds the number of data points. Moreover, the prior distribution is often chosen out of computational convenience instead of being motivated by domain knowledge.

\textit{Distance-aware methods} have a more pragmatic nature. They use the dissimilarity of a new input compared to the training data as a metric for model uncertainty \citep{lee2018simple, van2020uncertainty, ren2021simple}. As we will see in the next section, our method also exhibits this distance-aware property, albeit for a different reason: The further away from the training data, the easier it becomes for the network  to change the predictions without negatively affecting the likelihood of the training data.

\textit{Frequentist methods} use classical parametric statistics to obtain model uncertainty estimates. The typical approach - more elaborately explained in textbooks on parametric statistics, e.g. \cite{nonlinear2003} - involves obtaining (an estimate of) the variance of the model parameters using asymptotic theory and then converting this variance to the variance of the model predictions using the delta method. This approach has been used by various authors to create confidence intervals for neural networks \citep{kallus2022Implicit, nilsen2022Epistemic, deng2023Uncertainty, khosravi2011Comprehensive}.

Confidence intervals of this type are often referred to as Wald-type intervals. These intervals are necessarily symmetric. Various authors have noted that, for classical models, Wald-type intervals often behave worse than likelihood-ratio type intervals in the low-data regime \citep{hall1990Methodology, andersen2012Statistical, murphy1995Likelihood, murphy1997Semiparametric}. Specifically, when the loglikelihood cannot be effectively approximated with a quadratic function, Wald-type intervals may behave very poorly \citep[Chapter 2]{pawitan2001All}. The significant advantage of Wald-type intervals in the classical setting is the easier computation. However, while only a single model needs to be fitted, the necessary inversion of a high-dimensional $p \times p$ matrix and the quadratic approximation of the likelihood strongly rely on being in the asymptotic regime.

Conversely, the construction of the DeepLR confidence interval does not rely on a quadratic approximation of the likelihood, which is in general only valid asymptotically. While we still utilize asymptotic theory to determine the distribution of our test statistic (see our proof of Theorem \ref{thm: testdisttheorem} in Appendix \ref{prooftestthm}), we do impose the extremely strong requirement that the second derivative of the loglikelihood must converge and be invertible. We only require this second derivative to behave nicely in a single direction, which is a much weaker requirement. 

Another benefit of likelihood-ratio-based confidence intervals is that they are transformation invariant \citep{pawitan2000Reminder}. In other words, a different parametrization of the distribution does not alter the resulting confidence intervals. This is not the case for most Wald-type intervals or Bayesian approaches.

The main limitation of DeepLR in its current form is the computational cost. We compare the computational costs of the various approaches in Table \ref{computational costs}. Our method comes at no extra training cost but requires two additional networks to be trained for every confidence interval. Ensembling methods also need to train multiple networks, typically from five to ten, but these networks can be reused for different confidence intervals. Frequentist methods typically come at no additional training cost but require the inversion of a $p \times p$ matrix to construct the confidence interval. The cost of Bayesian methods varies drastically from method to method. The training process can be substantially more involved and the construction of a credible set requires multiple samples from the (approximate) posterior.
\begin{table}[tb]
\centering
\begin{tabularx}{\textwidth}{|X|X|X|}
\hline
Approach  & Additional training cost  & Additional inference cost \\
\hline
DeepLR  & No additional costs  & Two additional networks per new input and multiple additional forward passes. \\
Ensembling  & Multiple, typically five to ten, extra networks need to be trained. & A forward pass through each of the ensemble members. \\
Frequentist & None & Inversion of a $p\times p$ matrix. \\
Bayesian & Varies from method to method. MC dropout requires no additional training costs. & Varies from method to method. Typically, a large number of forward passes have to be made for every new input. \\

\hline
\end{tabularx}
\caption{A comparison of the computational costs of different types of methods that create confidence intervals or credible regions.}
\label{computational costs}
\end{table}

In summary, a likelihood-ratio-based method is distance aware, transformation invariant, has no accuracy loss, and is capable of creating asymmetric confidence intervals. However, it comes with the downside of being computationally expensive. Producing a confidence interval for a single input requires the training of two additional networks.

\section{Experimental results}\label{results}
In this section, we present the results of various experiments that showcase the desirable properties of DeepLR, such as its distance-aware nature and capability to create asymmetric confidence intervals. The high computational cost of our method prohibits any large-scale experiments. Nevertheless, the following experiments demonstrate the effectiveness of a likelihood-ratio-based approach.

All code used in the following experiments can be found at \url{https://github.com/LaurensSluyterman/Likelihood_ratio_intervals}
\subsection{Toy examples}
To start, we present two one-dimensional toy examples that effectively illustrate the behavior of our method. Specifically, these examples illustrate the capability of our method to produce asymmetric confidence intervals that expand in regions with a limited amount of data points, both during interpolation and extrapolation.

\paragraph{Regression}
The data set consists of 80 realisations of the random variable pair $(X,Y)$. Half of the $x$-values are sampled uniformly from the interval $[-1, -0.2]$, while the remaining half are sampled uniformly from the interval $[0.2, 1]$. The $y$-values are subsequently sampled using

\[
Y \mid X = x \sim \N{2x^{2}}{0.1^{2}}.
\]

On this training set, we train a mean-variance estimation network (MVE) \citep{nix1994estimating}. This particular type of network provides both mean and variance estimates and is trained by minimizing the negative loglikelihood under the assumption of a normal distribution. The network is trained for 400 epochs, using a default Adam optimizer \citep{kingma2014adam} and a batch size of 32. The network consists of 3 hidden layers for the mean estimation network with 40, 30, and 20 hidden units respectively and 2 hidden units for the variance using 5 and 2 hidden units respectively. All layers have elu activation functions \citep{clevert2015fast} with $l_{2}$-regularization applied in each dense layer with a constant value of 1e-4.

\begin{figure}[h!]
\centering
\begin{subfigure}{0.49\textwidth}
  \centering
  \includegraphics[width=\linewidth]{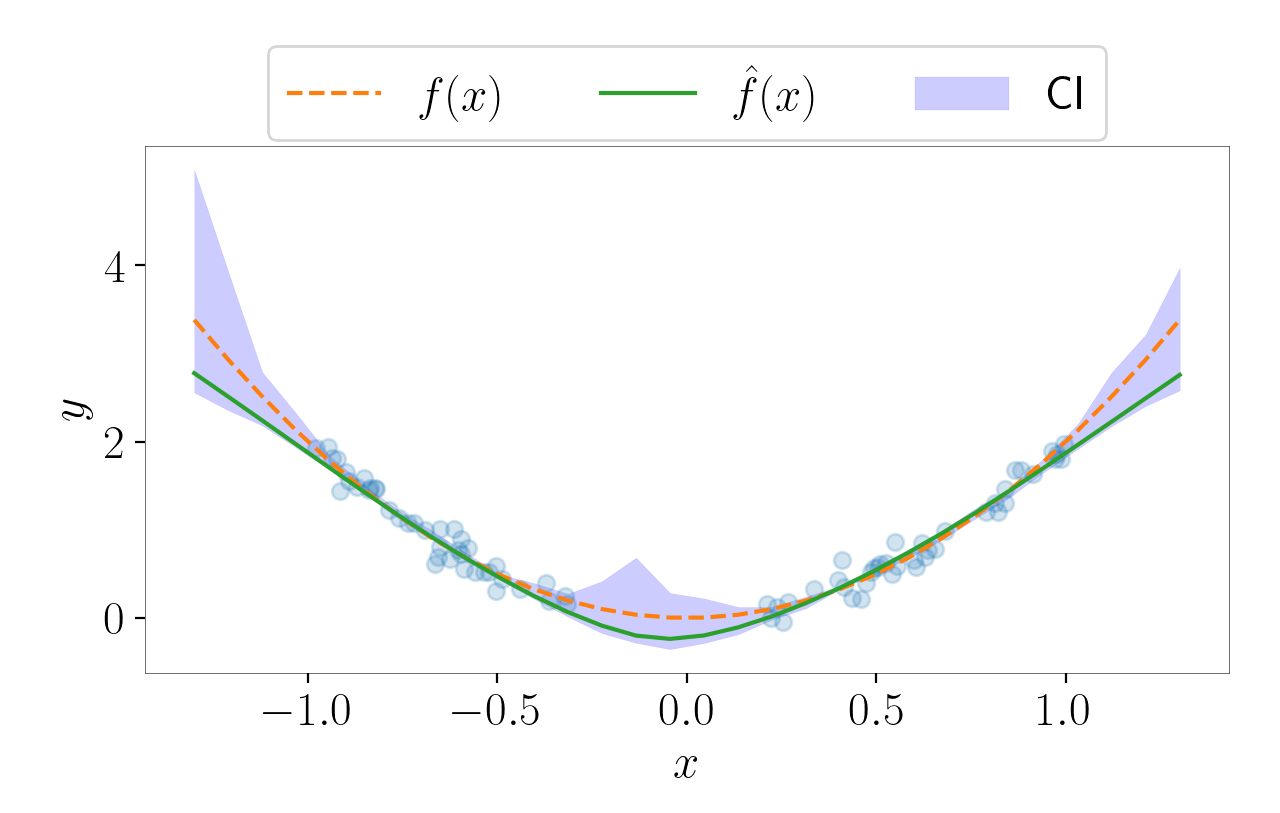}
  \caption{DeepLR}
  \label{fig:1}
\end{subfigure}
\hfill
\begin{subfigure}{0.49\textwidth}
  \centering
  \includegraphics[width=\linewidth]{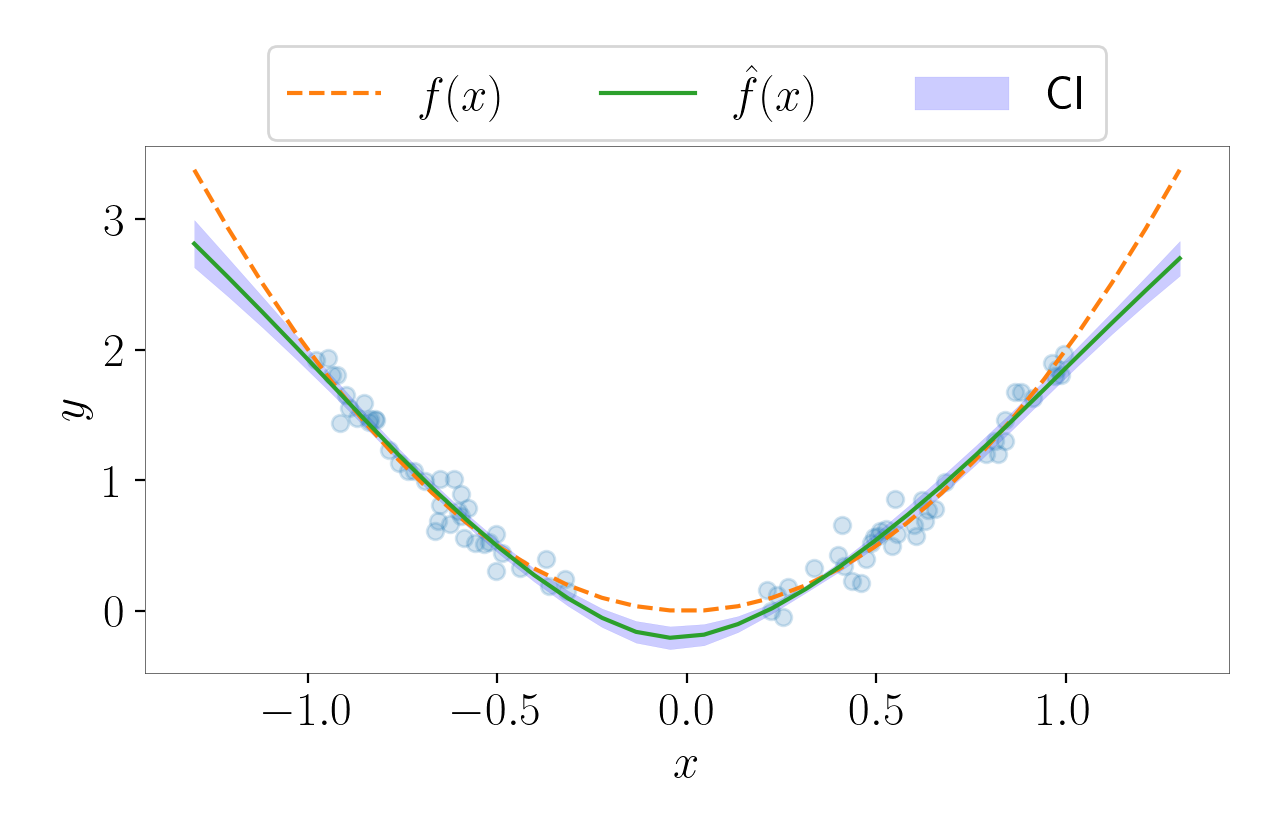}
  \caption{Ensemble}
  \label{fig:2}
\end{subfigure}

\caption{This figure illustrates DeepLR for a regression problem. The blue dots indicate the locations of the training points, the dotted orange line represents the true function, and the solid green line represents the predicted regression function of the network. The shaded blue region gives the 95$\%$ CI of the regression function. DeepLR exhibits two desirable properties when compared to an ensemble approach. Firstly, the intervals expand in regions where data is sparse. Secondly, the intervals can be asymmetric, allowing for the compensation of potential bias.}
\label{fig: toyreg}
\end{figure}

Figure \ref{fig: toyreg} gives the 95$\%$ confidence intervals for mean predictions. Notably, in the biased region around 0, the intervals become highly asymmetric. In contrast, most other methods produce symmetric interval around the original network. In a region with a bias, this can easily lead to intervals with very poor coverage. This is exemplified by the biased symmetric intervals generated by an ensemble consisting of 10 ensemble members (we used the ensembling strategy employed by \citet{lakshminarayanan2017simple}).


\paragraph{Binary classification}
The data set consists of 60 realisations of the random variable pair $(X, Y$), where half of the $x$-values are sampled uniformly from the interval $[0, 0.2]$ and the other half are sampled uniformly from the interval $[0.8, 1]$. The $y$-values are subsequently simulated using

\[Y \mid X =x \sim \text{Ber}(p(x)), \quad \text{with }p(x)=0.5 + 0.4\cos(6x). \]

On this training set, we train a fully connected network with three hidden layers consisting of 30 hidden units with elu activations functions. The final layer outputs a logit that is transformed using a sigmoid to yield a class probability. The network is trained for 300 epochs using a binary-crossy-entropy loss function and the Adam optimizer with a batch size of 32.

\begin{figure}[tb]
\centering
\begin{subfigure}{0.49\textwidth}
  \centering
  \includegraphics[width=\linewidth]{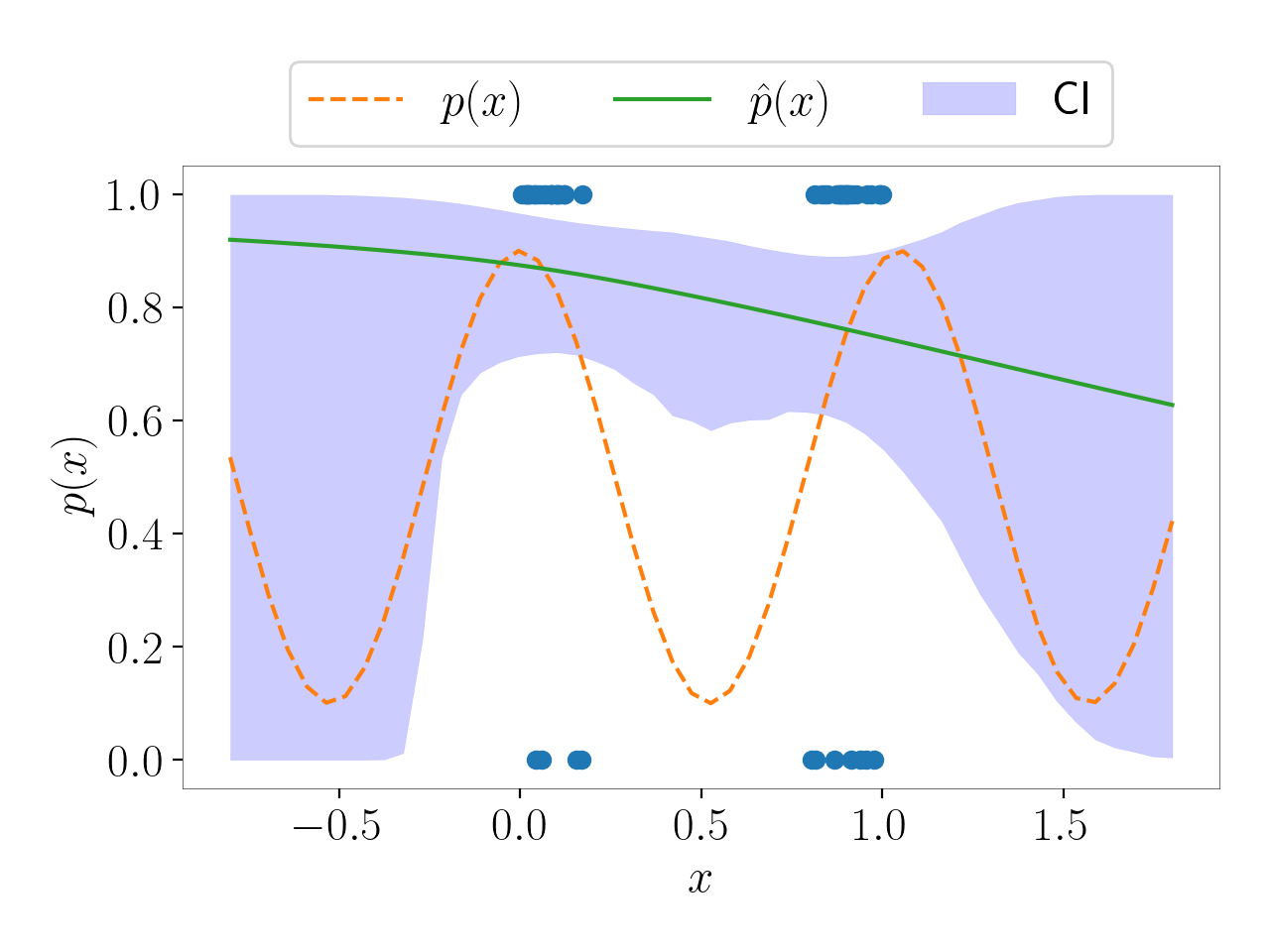}
  \caption{DeepLR, regularization constant 1e-5}
  \label{fig:1}
\end{subfigure}
\hfill
\begin{subfigure}{0.49\textwidth}
  \centering
  \includegraphics[width=\linewidth]{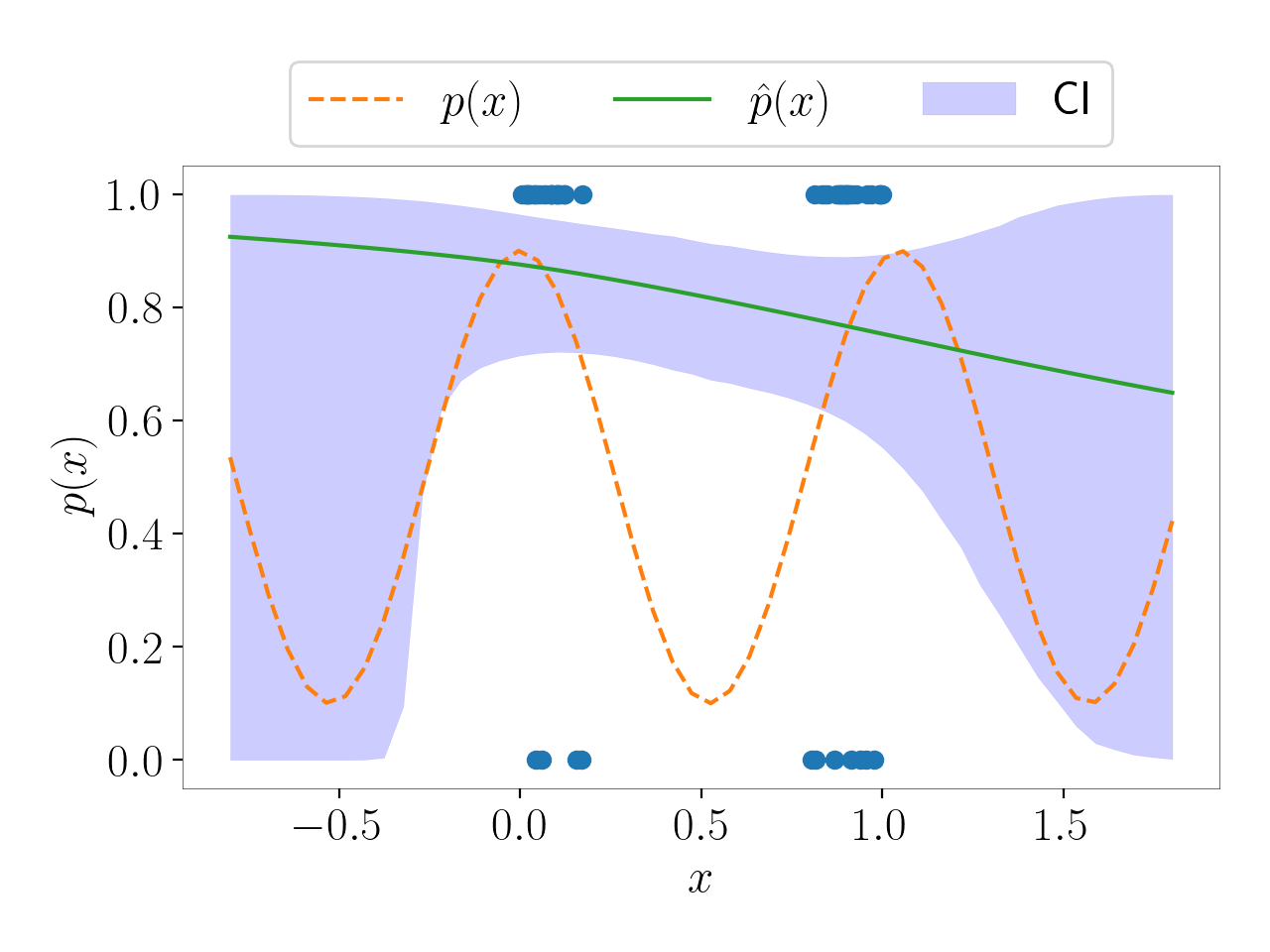}
  \caption{DeepLR, regularization constant 1e-4}
  \label{fig:2}
\end{subfigure}

\vspace{0.5cm}

\begin{subfigure}{0.49\textwidth}
  \centering
  \includegraphics[width=\linewidth]{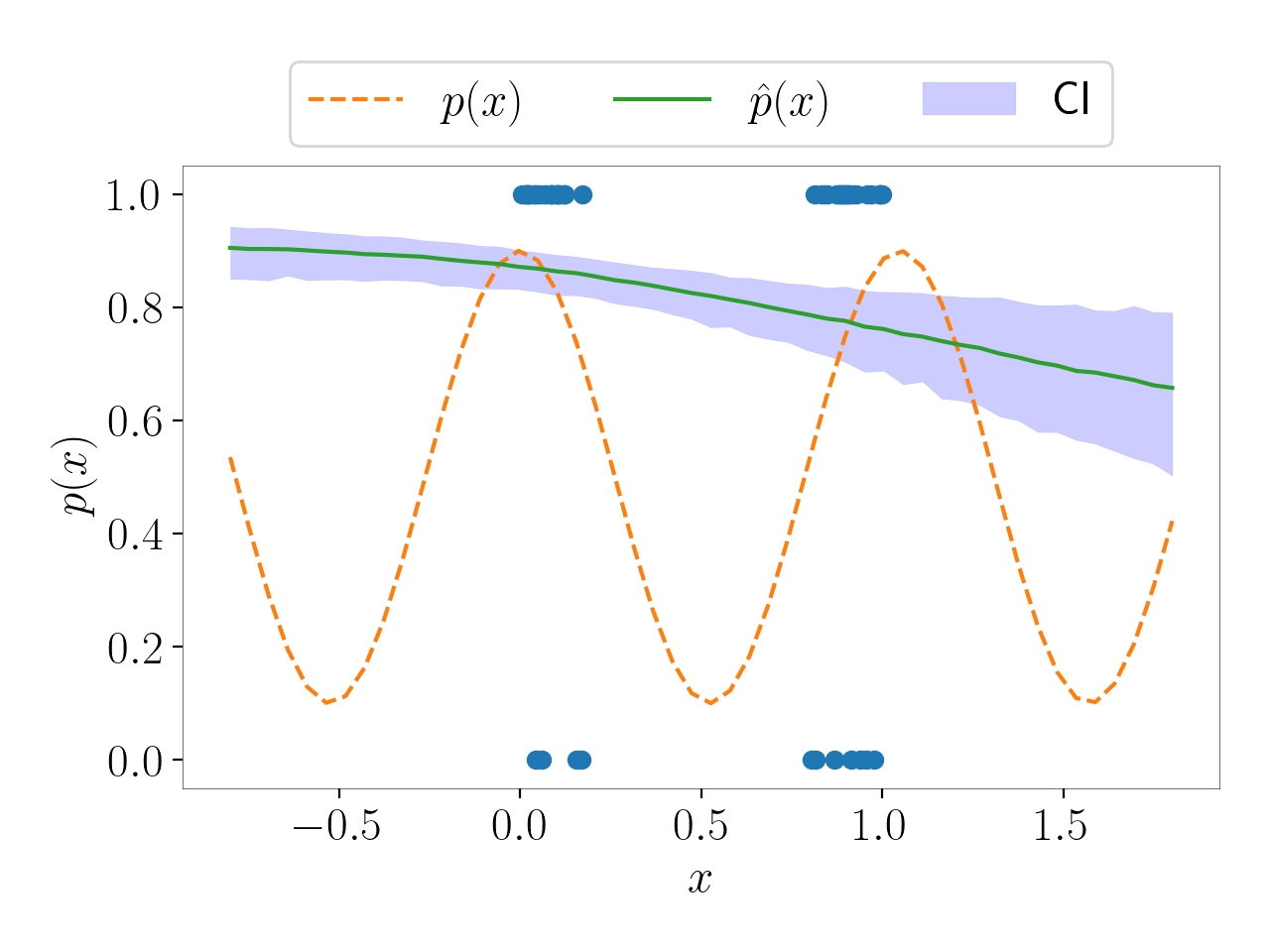}
  \caption{Dropout}
  \label{fig:3}
\end{subfigure}
\hfill
\begin{subfigure}{0.49\textwidth}
  \centering
  \includegraphics[width=\linewidth]{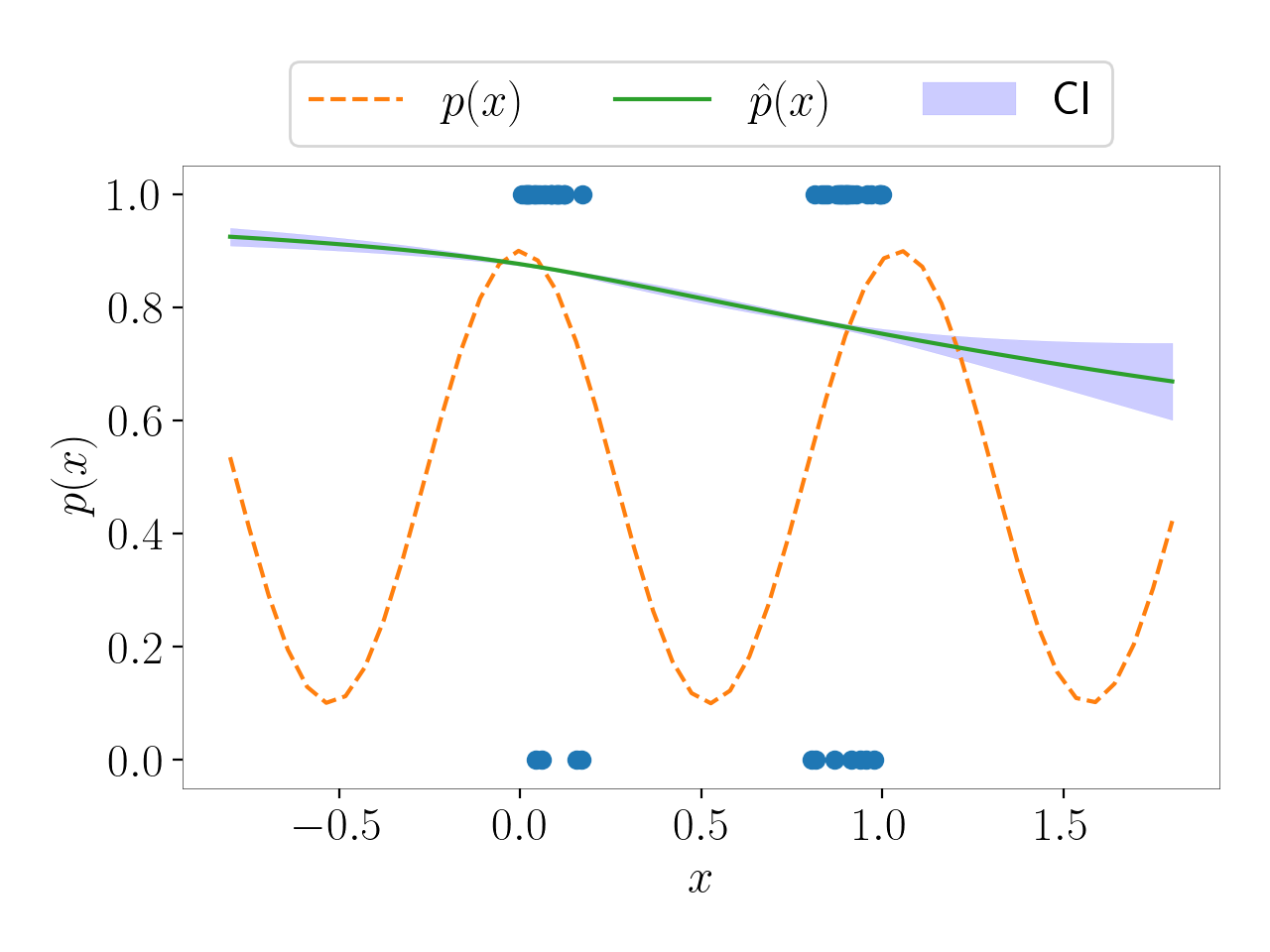}
  \caption{Ensemble}
  \label{fig:4}
\end{subfigure}
\caption{This figure illustrates DeepLR for a binary classification problem. The blue dots represent the training data, the dotted orange line represents the true probability of class 1 and the solid green line represents the predicted probability of class 1. The shaded blue region provides the 95$\%$ CI for the predicted probability of class 1. The top two figures (a) and (b) demonstrate the behavior of DeepLR for varying amounts of regularization. The more regularization, the smaller the class of admissible functions becomes, which naturally results in smaller intervals. Additionally, the top left figure demonstrates that the intervals expand when interpolating, a feature not shared by the dropout and ensembling approach.}
\label{fig: toyclass}
\end{figure}

The resulting 95$\%$ confidence intervals of the predicted probability of class 1 are given in Figure \ref{fig: toyclass}. We carried out the experiment with two amounts of regularization to illustrate how this affects the result. For comparison, we also implemented an ensembling approach and MC-Dropout (see \citet{lakshminarayanan2017simple} and \citet{gal2016dropout} for details). We used ten ensemble members and a standard dropout rate of 0.2. All networks were trained using the same training procedure.

We observe the same desirable properties as in the regression example. The intervals get much larger in regions with a limited amount of data, also when interpolating, and can become asymmetrical. Additionally, the intervals get smaller when we increase the amount of regularization. The model class becomes smaller (fewer parameters can be reached) making it more difficult for the model to change the predictions without affecting the likelihood. This, in turn, leads to smaller confidence intervals. If the model is overly regularized, it will become miss specified ($\theta_{0} \notin \Theta$) and the resulting confidence intervals will not be correct.

The other approaches do not share the same qualitative properties. The ensembling approach results in confidence intervals that are far too narrow. All ensemble members behave more or less the same, especially when interpolating. The MC-Dropout credible regions do not expand in the regions in between the data and also only moderately expand when extrapolating.

\subsection{Two-moon example}
The data set consists of 80 data points, generated using the \texttt{make\_moons} function from the \texttt{scikit-learn} package, which creates a binary classification problem with two interleaving half circles \citep{pedregosa2011Scikitlearn}. 

We utilize the same network architecture as in the toy classification example. The network is trained for 500 epochs using the Adam optimizer with default learning rate and batch size of 32, while also applying $l_2$-regularization in each layer with a constant value of 1e-3.

Figure \ref{fig: twomoon} presents the 95$\%$ confidence intervals for the predicted class probabilities. The results illustrate that DeepLR becomes extremely uncertain in regions farther from the data (i.e., the confidence interval for the class probability spans the full range of $[0, 1]$). 

In contrast, both an ensembling approach and MC-Dropout report excessively high certainty in the upper left and lower right regions. The ensemble's behavior can be attributed to all ensemble members extrapolating in the same direction causing all ensemble members to report more or less the same class probability. For MC-dropout, a saturated sigmoid is causing the narrow credible intervals. 

This comparison underscores the unique capability of DeepLR to provide more accurate uncertainty estimates in regions less well represented by data -- a crucial capability in practical applications.
\subsection{MNIST binary example}
For a more difficult task, we train a small convolutional network on the first two classes of the MNIST data set, consisting of handwritten digits. In this binary classification task, the 0's labeled as class 0 and the 1's as class 1. 

The CNN architecture consists of two pairs of convolutional layers (with 28 filters and 3x3 kernels) and max-pooling layers (2x2 kernel), followed by a densely connected network with two hidden layers with 30 hidden units each and elu activation functions.

 The network is trained for 10 epochs, using the SGD optimizer with a batch size of 32, default learning rate, and $l_2$ regularization with a constant value of 1e-5, and binary cross-entropy loss function. The amounts of training time and regularization were determined from a manual grid search using an 80/20 split of training data. For the actual experiment, the entire training set was utilized. 

\begin{figure}[h!]
\centering
\vskip -0.in
  \begin{minipage}{0.3\textwidth}
    \centering
   \textbf{Training}
  \end{minipage}
  \hfill
  \begin{minipage}{0.3\textwidth}
    \centering
	\textbf{Test}
  \end{minipage}
  \hfill
  \begin{minipage}{0.3\textwidth}
    \centering
	\textbf{OoD}
  \end{minipage}
   \vskip -0.0in
\begin{subfigure}{0.3\textwidth}
  \centering
  \includegraphics[width=\linewidth]{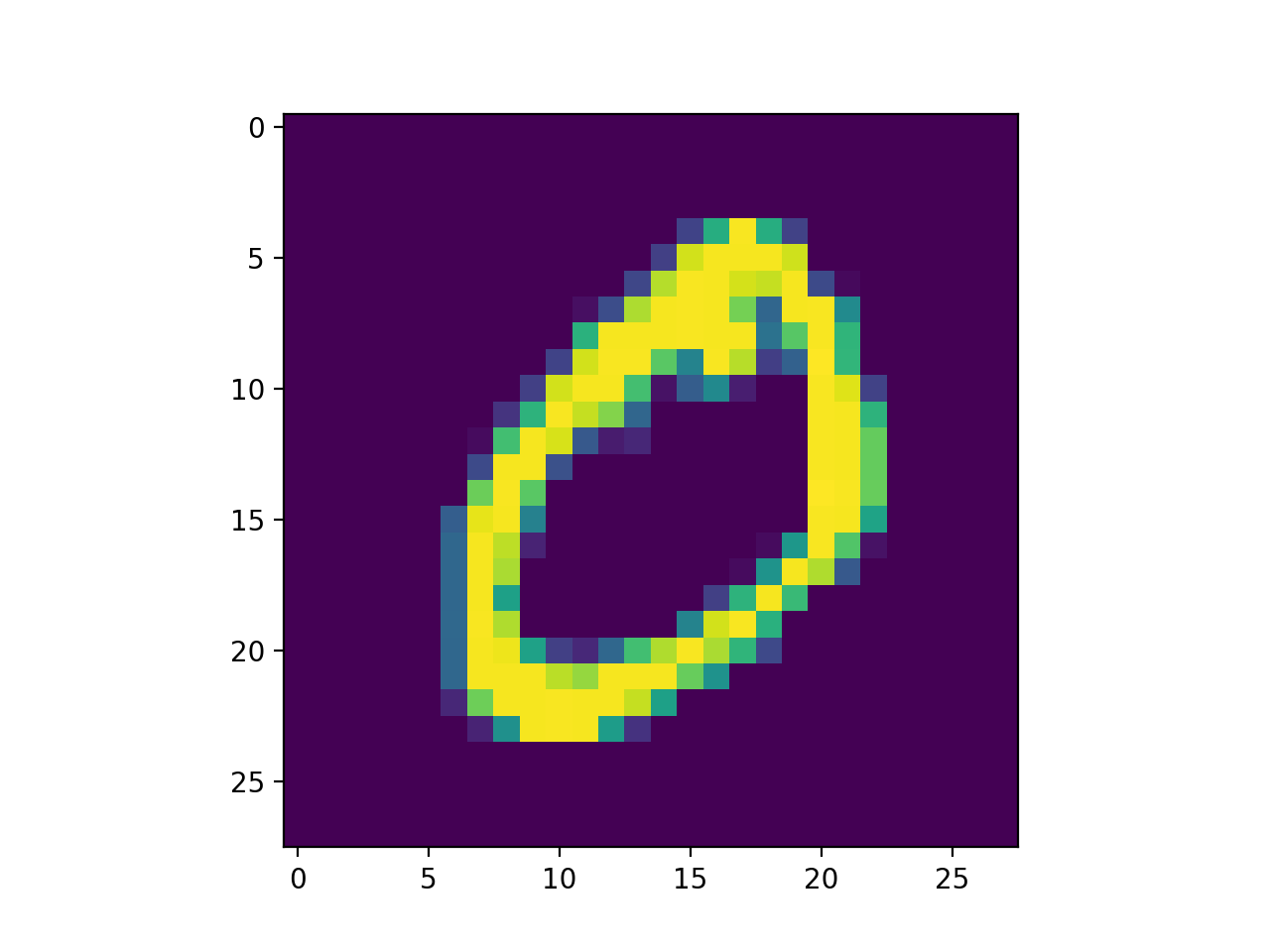}
  \vskip -0.in
  \caption{LR: [0.00, 0.00] \\ DR: [0.00, 0.00] \\ EN: [0.00, 0.00]}
  \label{fig:1}
\end{subfigure}
\hfill
\begin{subfigure}{0.3\textwidth}
  \centering
  \includegraphics[width=\linewidth]{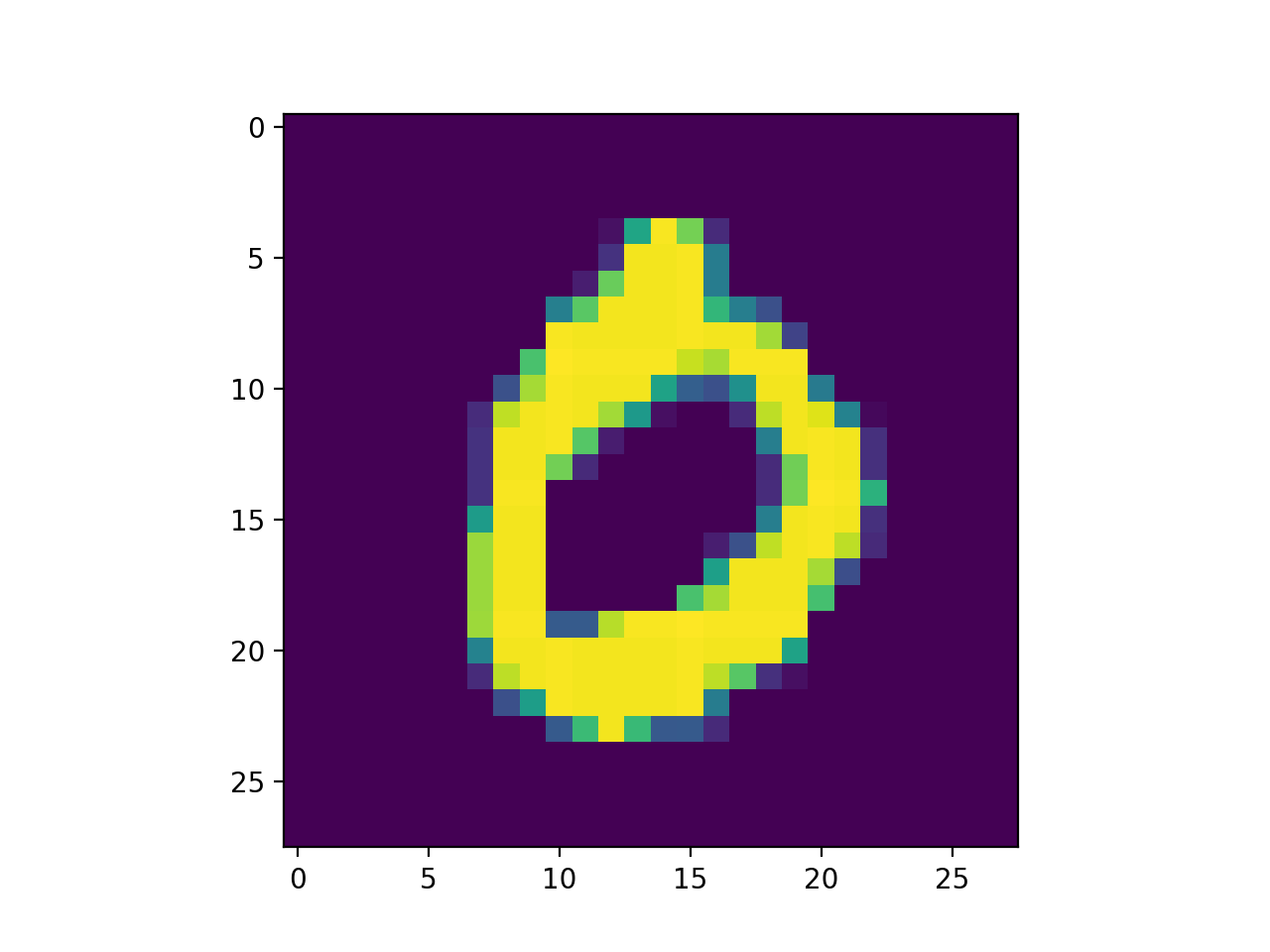}
  \vskip -0.in
  \caption{LR: [0.00, 0.00] \\ DR: [0.00, 0.00] \\ EN: [0.00, 0.00]}
  \label{fig:1}
\end{subfigure}
\hfill
\begin{subfigure}{0.3\textwidth}
  \centering
  \includegraphics[width=\linewidth]{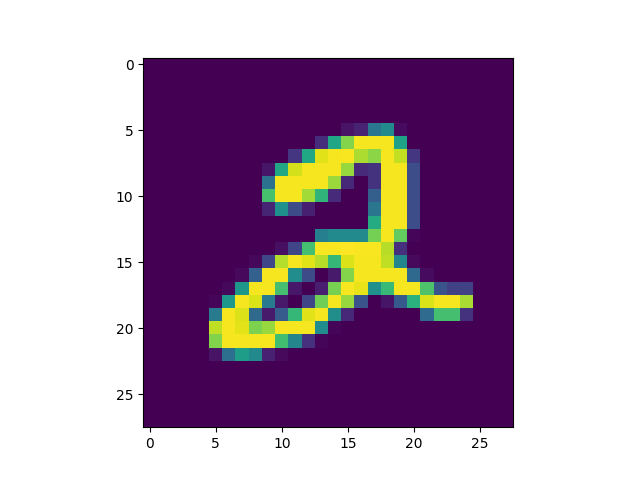}
  \vskip -0.in
  \caption{LR: [0.00, 0.13] \\ DR: [0.00, 0.07] \\ EN: [0.00, 0.10]}
 \end{subfigure}
 \vskip -0.in
\begin{subfigure}{0.3\textwidth}
  \centering
  \includegraphics[width=\linewidth]{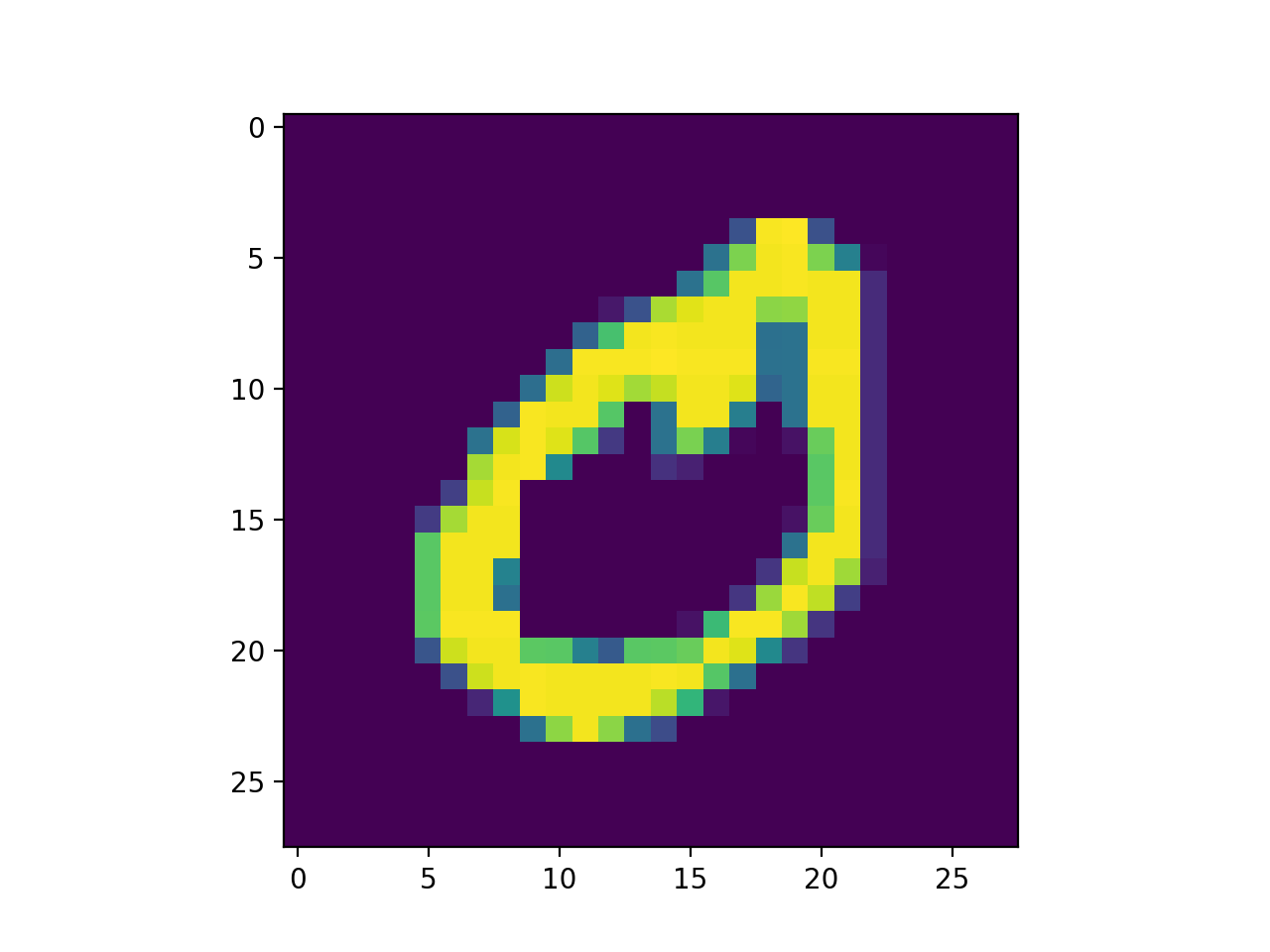}
  \vskip -0.in
  \caption{LR: [0.00, 0.00] \\ DR: [0.00, 0.00] \\ EN: [0.00, 0.00]} 
  \end{subfigure} \hfill
\begin{subfigure}{0.3\textwidth}
  \centering
  \includegraphics[width=\linewidth]{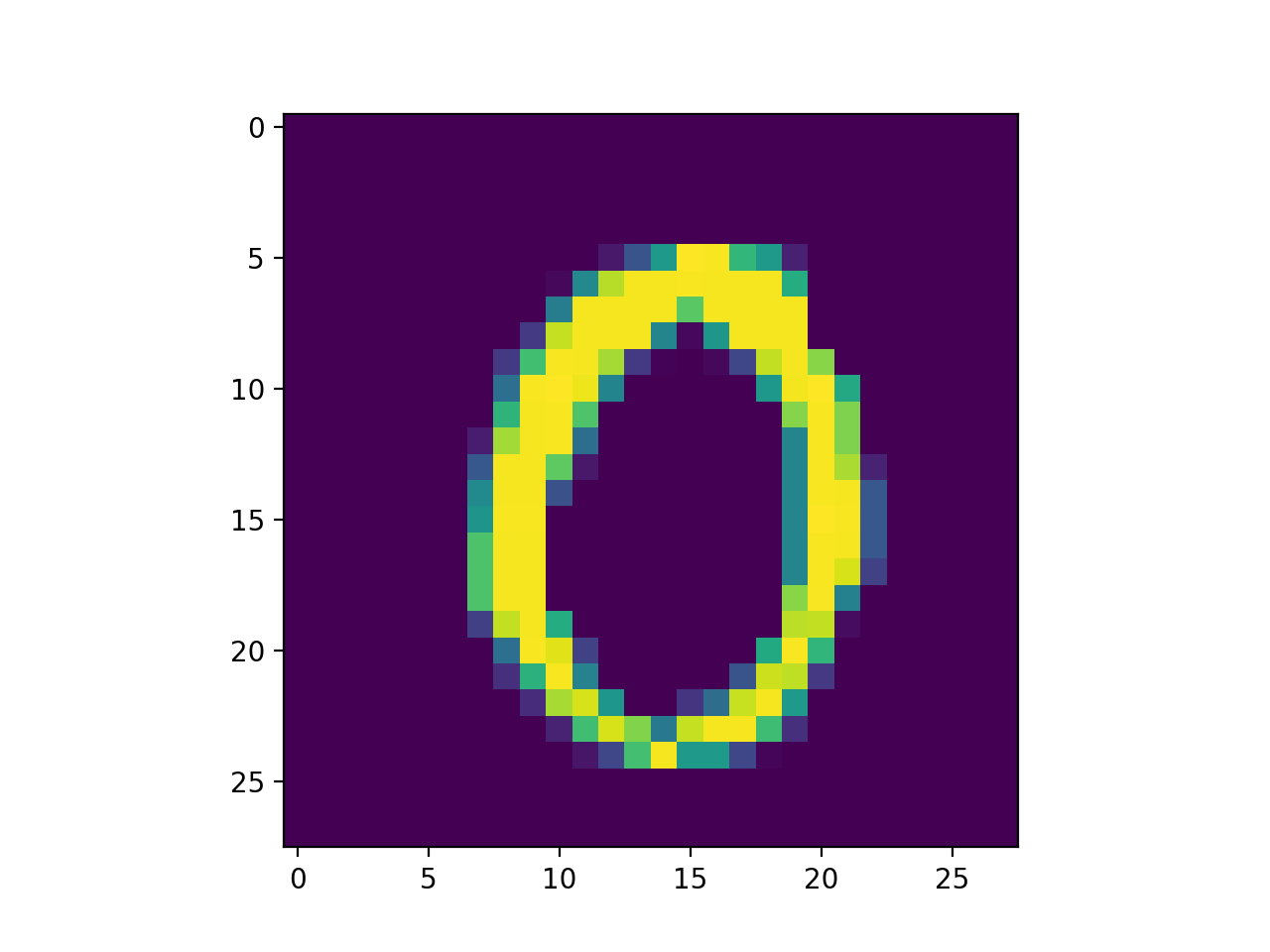}
  \vskip -0.in
  \caption{LR: [0.00, 0.00] \\ DR: [0.00, 0.00] \\ EN: [0.00, 0.00]}
  \label{fig:1}
\end{subfigure}
\hfill
\begin{subfigure}{0.3\textwidth}
  \centering
  \includegraphics[width=\linewidth]{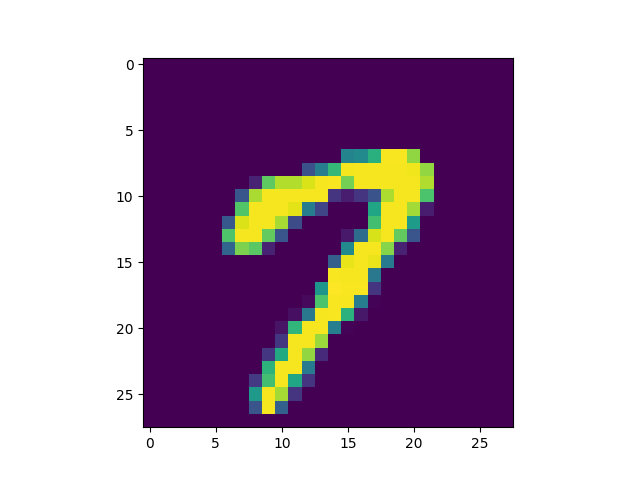}
  \vskip -0.in
  \caption{LR: [0.09, 0.39] \\ DR: [0.00, 0.10] \\ EN: [0.00, 0.21]}
\end{subfigure} 
 \vskip -0.in
\begin{subfigure}{0.3\textwidth}
  \centering
  \includegraphics[width=\linewidth]{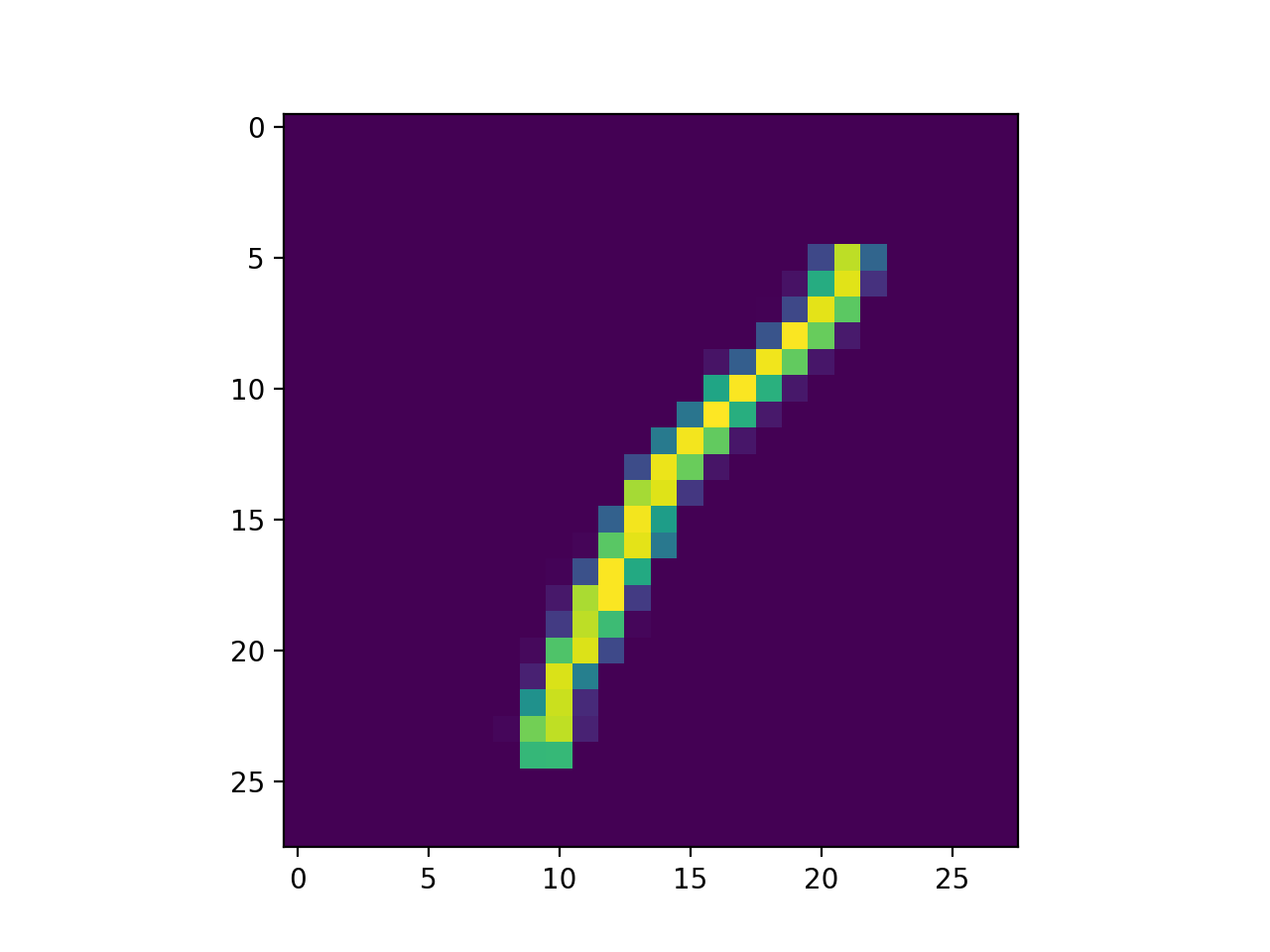}
  \vskip -0in
  \caption{LR: [1.00, 1.00] \\ DR: [0.99, 1.00] \\ EN: [1.00, 1.00] }
\end{subfigure}
\hfill
\begin{subfigure}{0.3\textwidth}
  \centering
  \includegraphics[width=\linewidth]{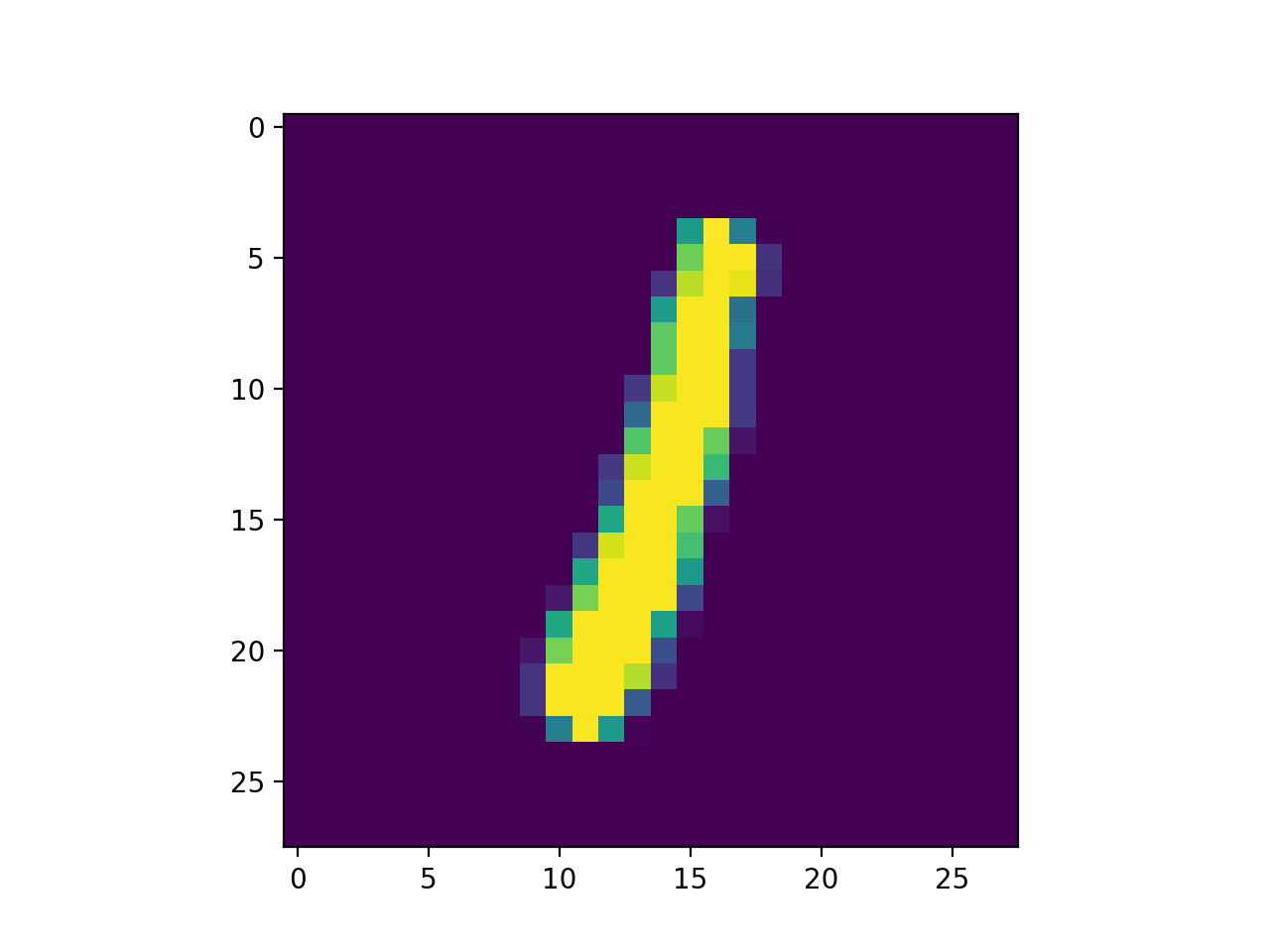}
  \vskip -0.in
  \caption{LR: [1.00, 1.00] \\ DR: [1.00, 1.00] \\ EN: [1.00, 1.00]}
  \label{fig:1}
\end{subfigure}
\hfill
\begin{subfigure}{0.3\textwidth}
  \centering
  \includegraphics[width=\linewidth]{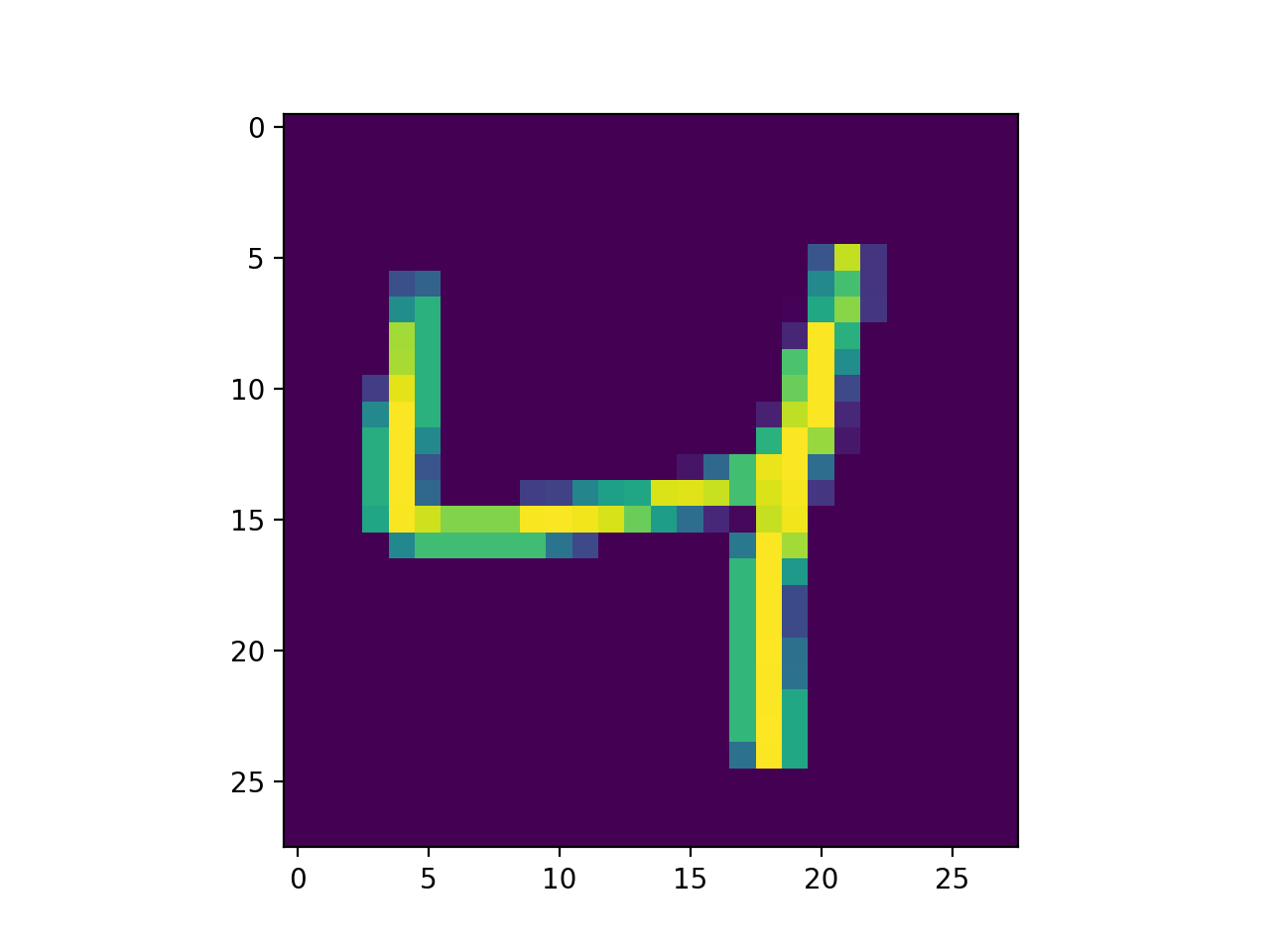}
  \vskip -0.in
  \caption{LR: [0.04, 0.33] \\ DR: [0.00, 0.03] \\ EN: [0.00, 0.04]}
\end{subfigure} 
\caption{95$\%$ CIs for different training points (first column), test points (second column), and OoD points (third column). Each subcaption displays the $95\%$ CI for the probability of class 1 made with DeepLR (LR), MC-dropout (DR), and ensembling (EN). For in-distribution points (zeros and ones), all three methods provide extremely narrow intervals. For the OoD points, our method provides much larger CIs, a property also present in the other methods, albeit to a lesser extent.}
\label{fig: mnist}
\end{figure}

Figure \ref{fig: mnist} presents 95$\%$ CIs for a number of different training points, test points, and OoD points. As shown, the OoD points have wider confidence intervals than the training and test points, reflecting greater uncertainty. 

\begin{figure}[h!]
\centering 
\vskip-0.1in
\begin{subfigure}{1\textwidth}
    \includegraphics[width=0.9\linewidth]{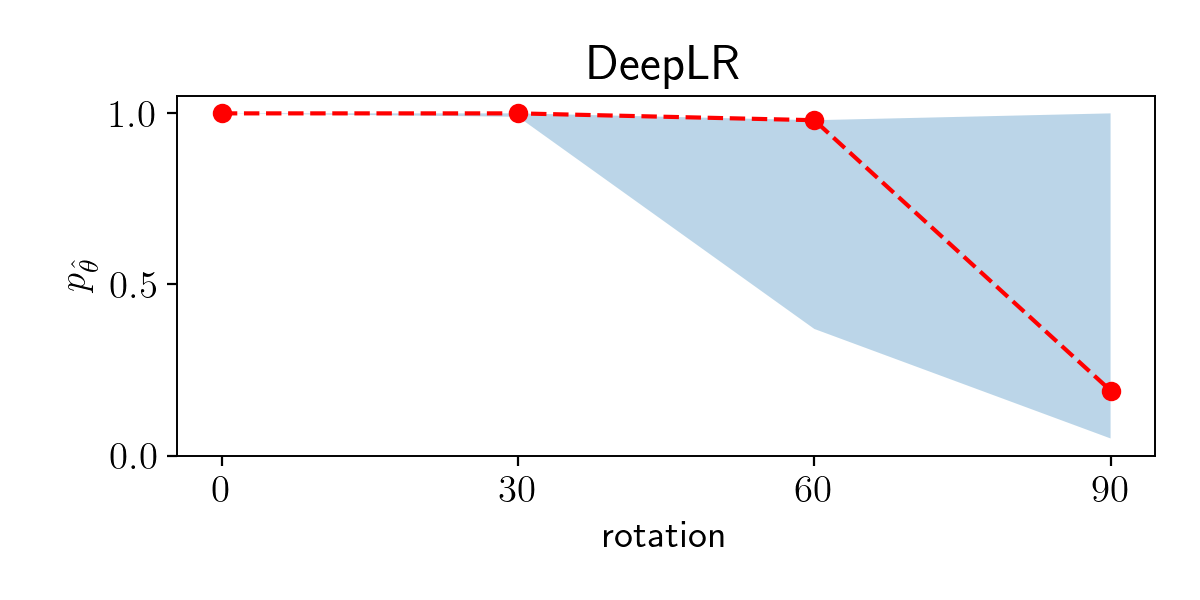} 
\end{subfigure}
\vskip-0.2in
\begin{subfigure}{1\textwidth}
    \includegraphics[width=0.9\linewidth]{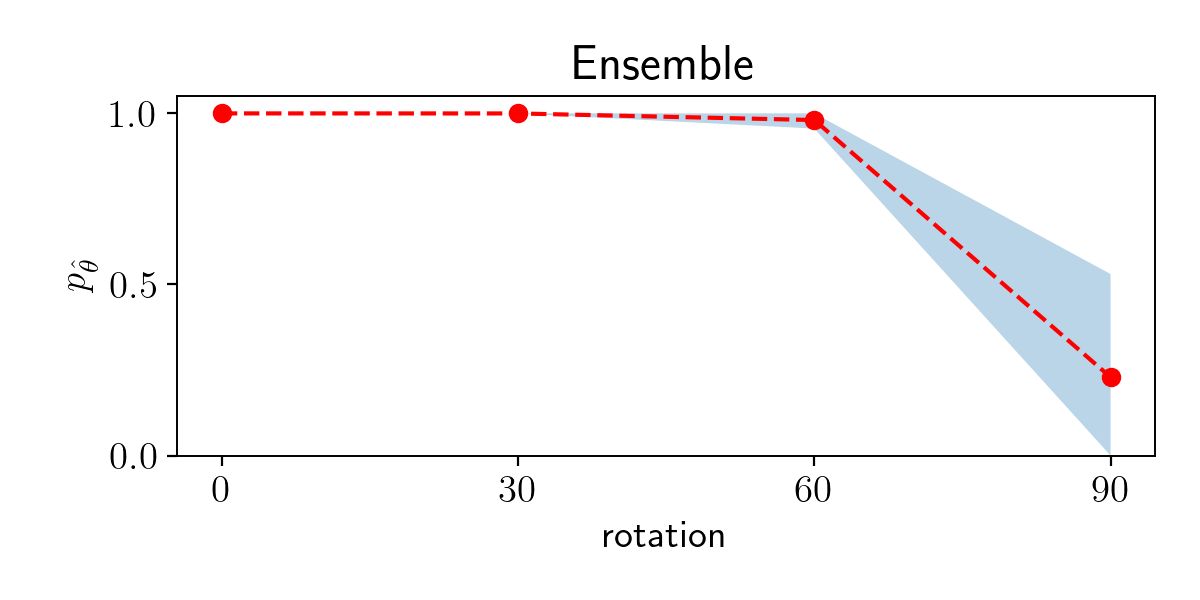} 
\end{subfigure}
\vskip -0.2 in
\begin{subfigure}{1\textwidth}
    \includegraphics[width=0.9\linewidth]{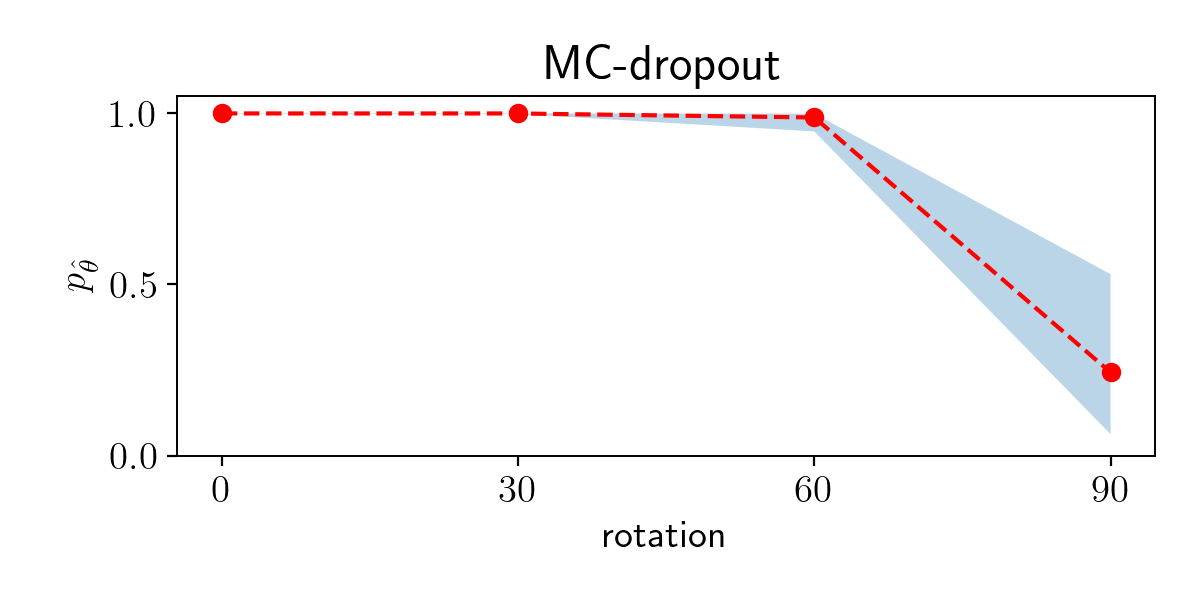} 
\end{subfigure}
\vskip -0.2 in
\begin{subfigure}{0.21\textwidth}
  \centering
 \includegraphics[width=0.9\linewidth]{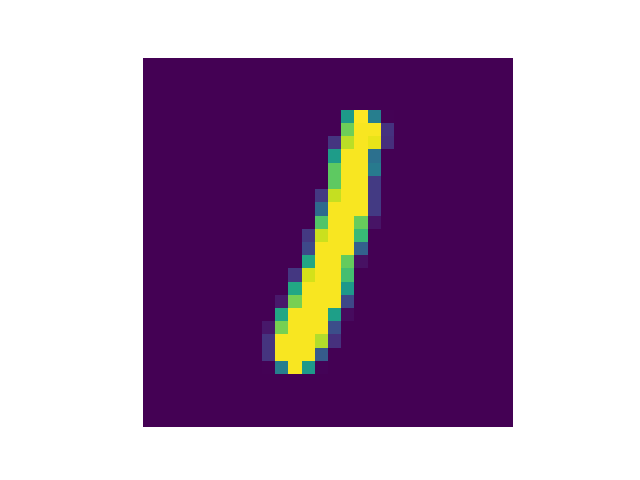}
  \caption*{$0^\circ$}
\end{subfigure}
\hskip 0.in
\begin{subfigure}{0.21\textwidth}
  \centering
  \includegraphics[width=0.9\linewidth]{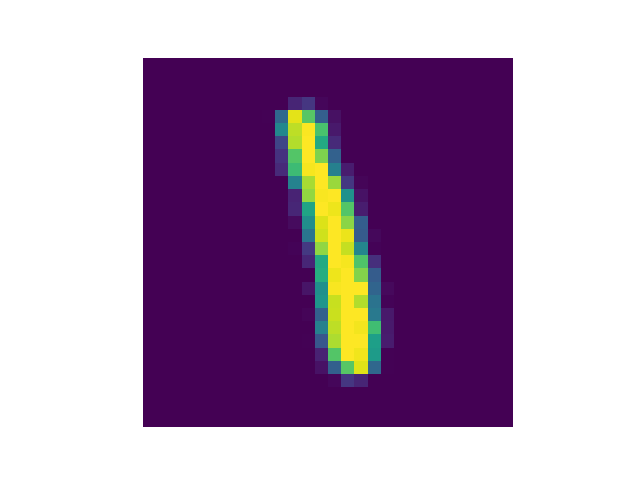}
  \caption*{$30^\circ$}
  \label{fig:1}
\end{subfigure}
\hskip 0.in
\begin{subfigure}{0.21\textwidth}
  \centering
  \includegraphics[width=0.9\linewidth]{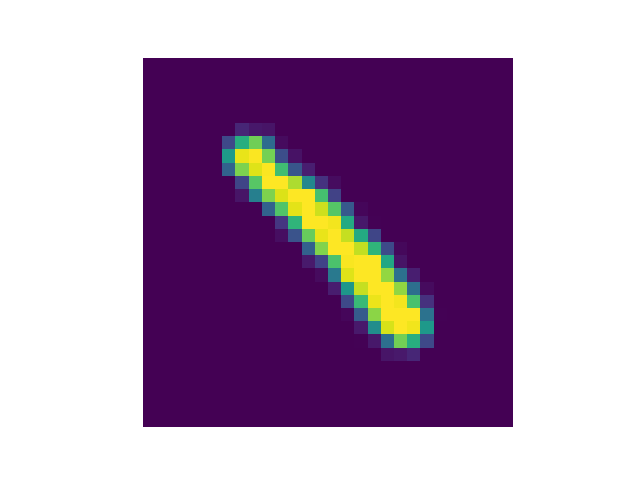}
  \caption*{$60^\circ$}
\end{subfigure} 
\hskip 0in
\begin{subfigure}{0.21\textwidth}
  \centering
  \includegraphics[width=0.9\linewidth]{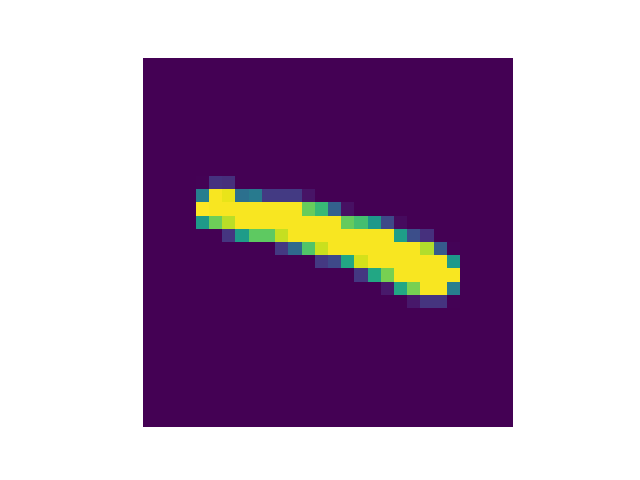}
  \caption*{$90^\circ$}
\end{subfigure}
\vskip -0.05 in
\caption{This figure provides 95$\%$ CIs for different amounts of rotation. For rotations of 0 and 30 degrees, all methods produce very narrow confidence intervals, implying high certainty. At 60 degrees of rotation, DeepLR outputs high uncertainty whereas the ensembling approach and MC-dropout remain fairly certain. At 90 degrees of rotation, DeepLR outputs very high uncertainty, a confidence interval of [0.05, 1.00], a behavior also observed to a lesser extent in both ensembling and MC-dropout.}
\vskip -0.05 in
\label{fig: mnistrot}
\end{figure}

In an additional experiment, we rotated one of the test-points and created confidence intervals for the rotated images. As Figure \ref{fig: mnistrot} illustrates, increasing the rotation angle results in larger confidence intervals. This behavior is also seen with the ensembling and MC-dropout, but to a significantly lesser extent. 

These larger intervals can be explained as follows: Our approach essentially asks the intuitive question how much the network can change the prediction for this new input without overly affecting the likelihood of the training data. A heavily rotated image of a number 1 deviates greatly from the typical input, utilizing pixels that are almost never used by the training inputs. It is therefore relatively straightforward for the network to change the prediction of this rotated input without dramatically lowering the likelihood of the training data. This, in turn, results in very large confidence intervals, accurately indicating that it is a very unfamiliar input.

\FloatBarrier
\subsection{CIFAR binary example}
We extend the previous experiment to the CIFAR10 data set, using the first two classes -- planes and cars -- as a binary classification problem. 

We use a CNN consisting of two pairs of convolutional layers (with 32 filters and 3x3 kernels) and max-pooling layers (2x2 kernel), followed by a densely connected network with three hidden layers with 30 hidden units each and elu activation functions. 

The CNN is trained for 15 epochs using the SGD optimizer with default learning rate, a batch size of 32, and $l_2$-regularization with a constant value of 1e-5. The training time and regularization are determined the same way as for the MNIST experiment, using an 80/20 split of the training data. 

\begin{figure}[h!]
\centering
  \begin{minipage}{0.3\textwidth}
    \centering
   \textbf{Training}
  \end{minipage}
  \hfill
  \begin{minipage}{0.3\textwidth}
    \centering
	\textbf{Test}
  \end{minipage}
  \hfill
  \begin{minipage}{0.3\textwidth}
    \centering
	\textbf{OoD}
  \end{minipage}
\begin{subfigure}{0.3\textwidth}
  \centering
  \includegraphics[width=\linewidth]{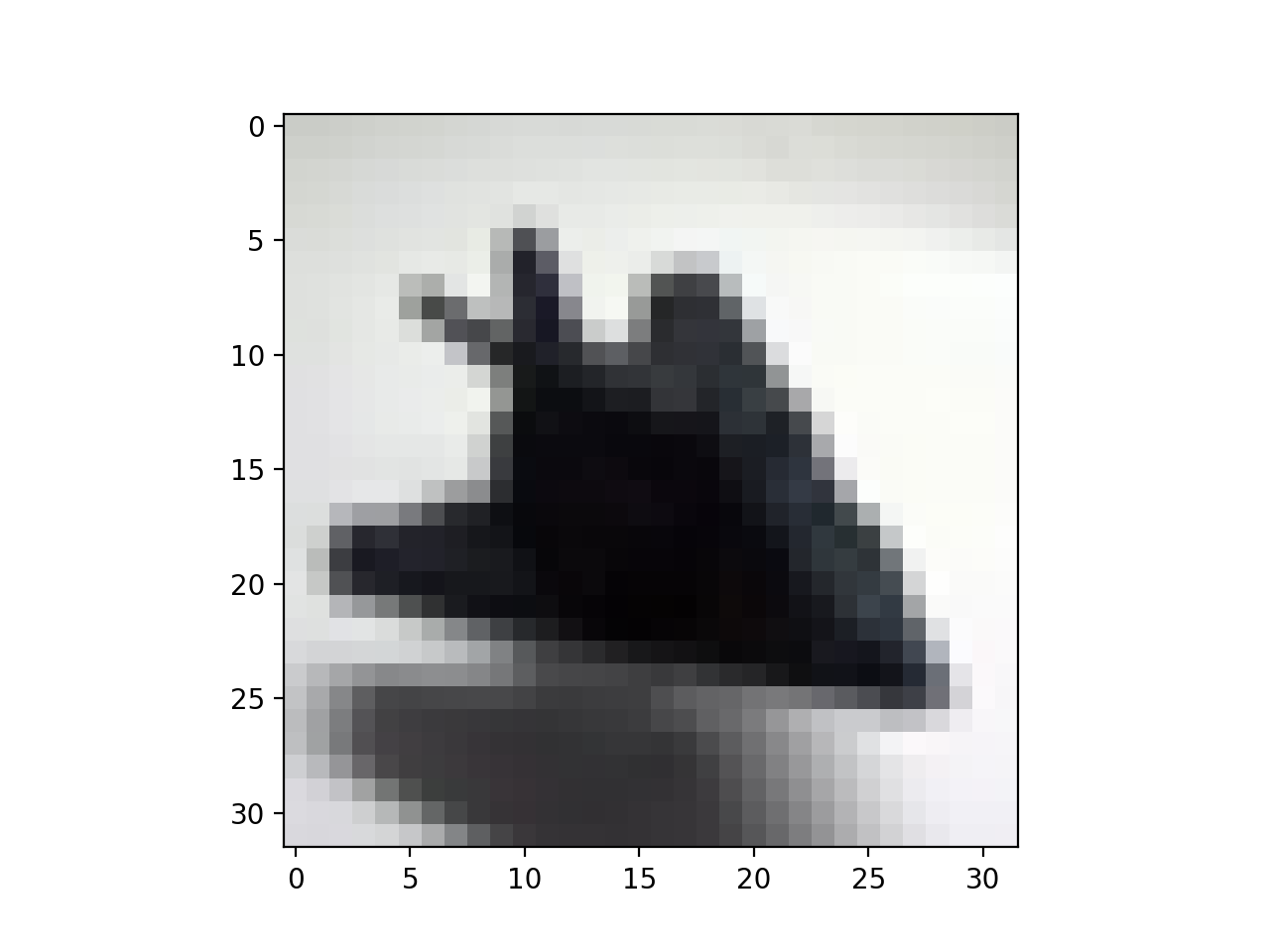}
  \vskip -0.in
  \caption{LR: [0.01, 0.30] \\ DR: [0.01, 0.34] \\ EN: [0.00, 0.20]}
\end{subfigure}
\hfill
\begin{subfigure}{0.3\textwidth}
  \centering
  \includegraphics[width=\linewidth]{CIFARtest0.png}
  \vskip -0.in
  \caption{LR: [0.01, 0.14] \\ DR: [0.00, 0.05] \\ EN: [0.00, 0.23]}
  \label{fig:1}
\end{subfigure}
\hfill
\begin{subfigure}{0.3\textwidth}
  \centering
  \includegraphics[width=\linewidth]{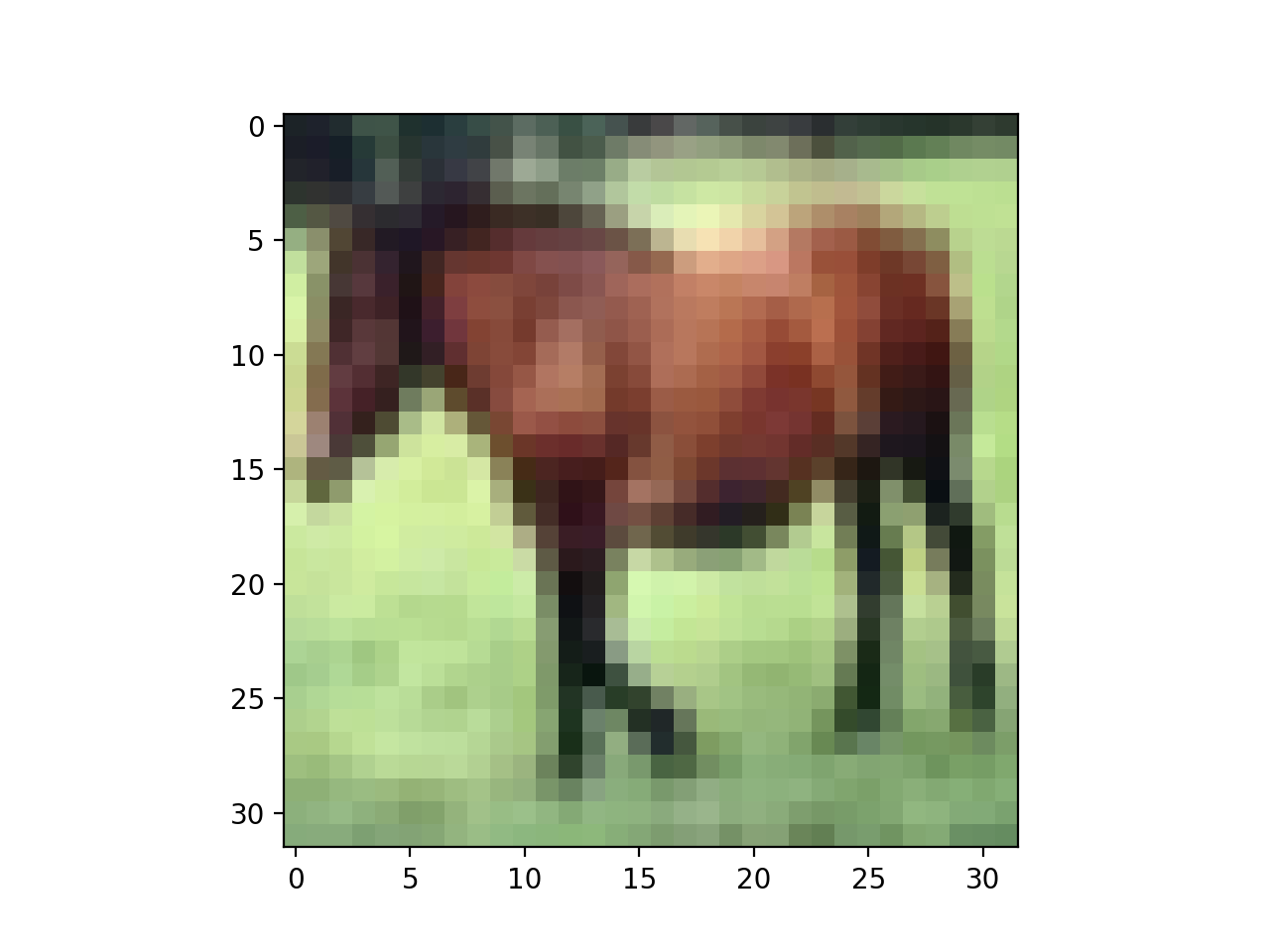}
  \vskip -0.in
  \caption{LR: [0.08, 0.94] \\ DR: [0.71, 0.97] \\ EN: [0.17, 1.00]]}
\end{subfigure} 
\vskip -0.in
\begin{subfigure}{0.3\textwidth}
  \centering
  \includegraphics[width=\linewidth]{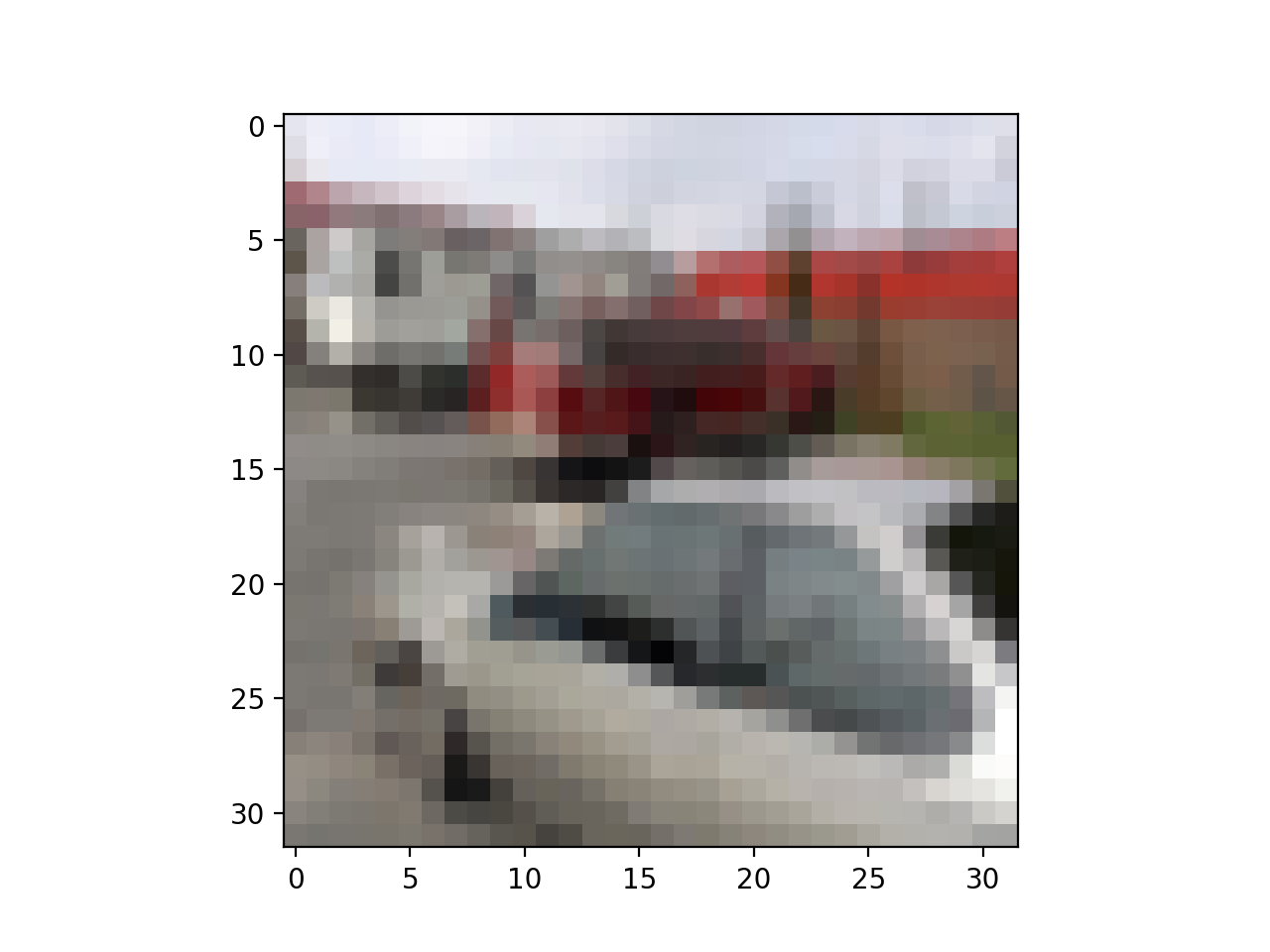}
  \vskip -0.in
  \caption{LR: [0.57, 0.95] \\ DR: [0.12, 0.57] \\ EN: [0.33, 1.00]}
\end{subfigure}
\hfill
\begin{subfigure}{0.3\textwidth}
  \centering
  \includegraphics[width=\linewidth]{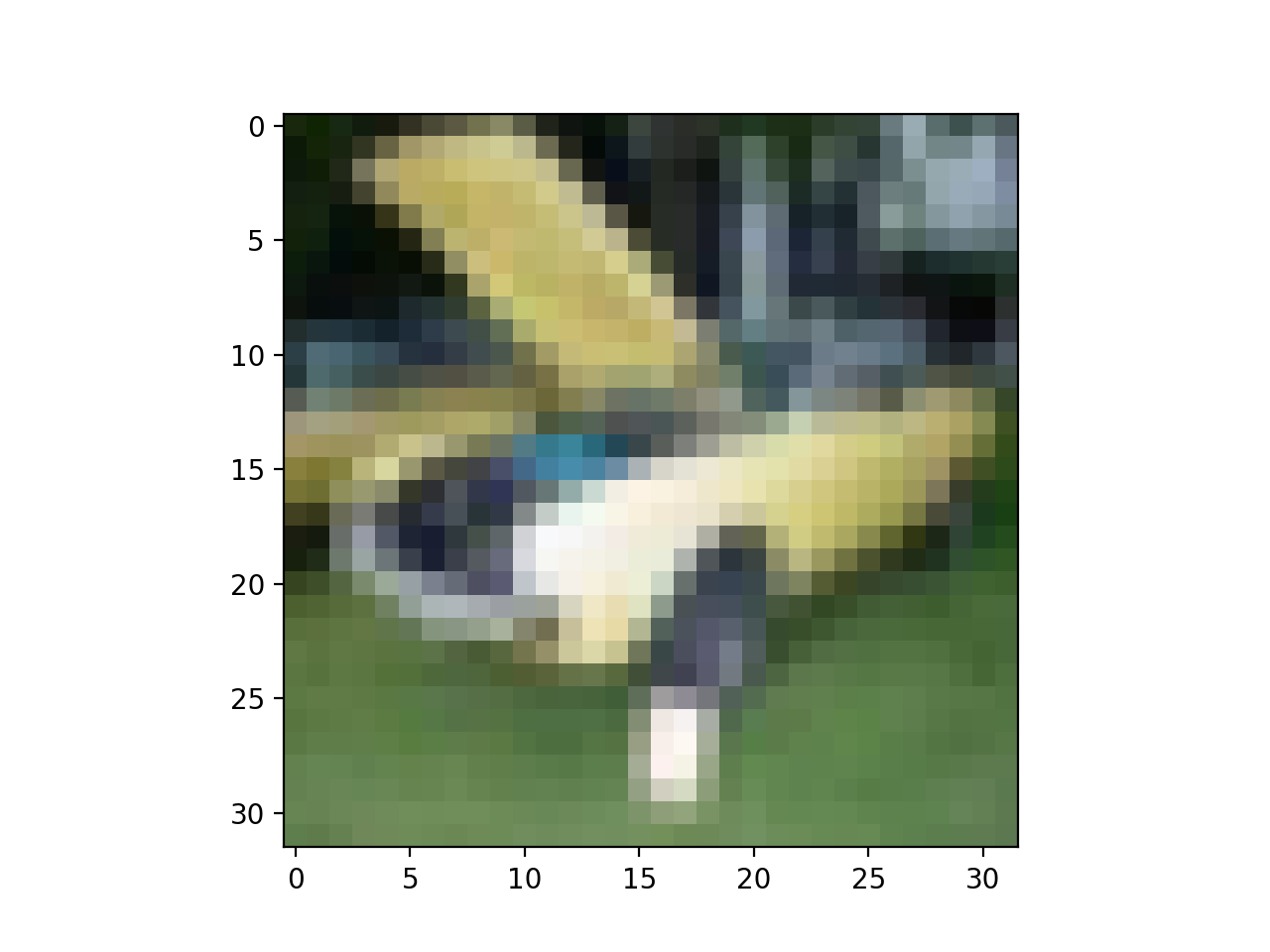}
  \vskip -0.in
  \caption{LR: [0.00, 0.93] \\ DR: [0.43, 0.91] \\ EN: [0.38, 1.00]}
  \label{fig:cifartest-1}
\end{subfigure}
\hfill
\begin{subfigure}{0.3\textwidth}
  \centering
  \includegraphics[width=\linewidth]{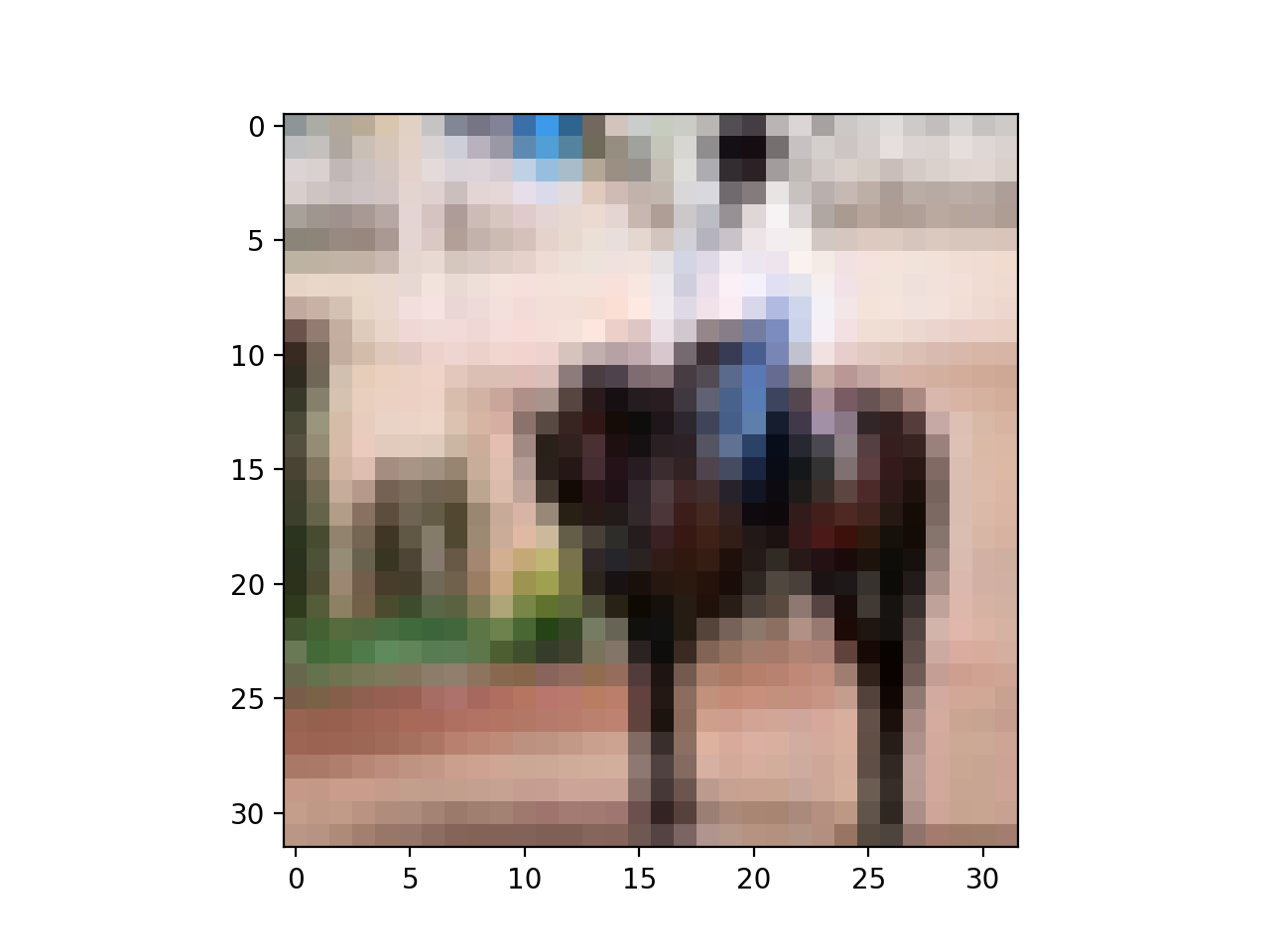}
  \vskip -0.in
  \caption{LR: [0.00, 1.00] \\ DR: [0.16, 0.69] \\ EN: [0.33, 1.00]}
\end{subfigure} 
\vskip -0.in
\begin{subfigure}{0.3\textwidth}
  \centering
  \includegraphics[width=\linewidth]{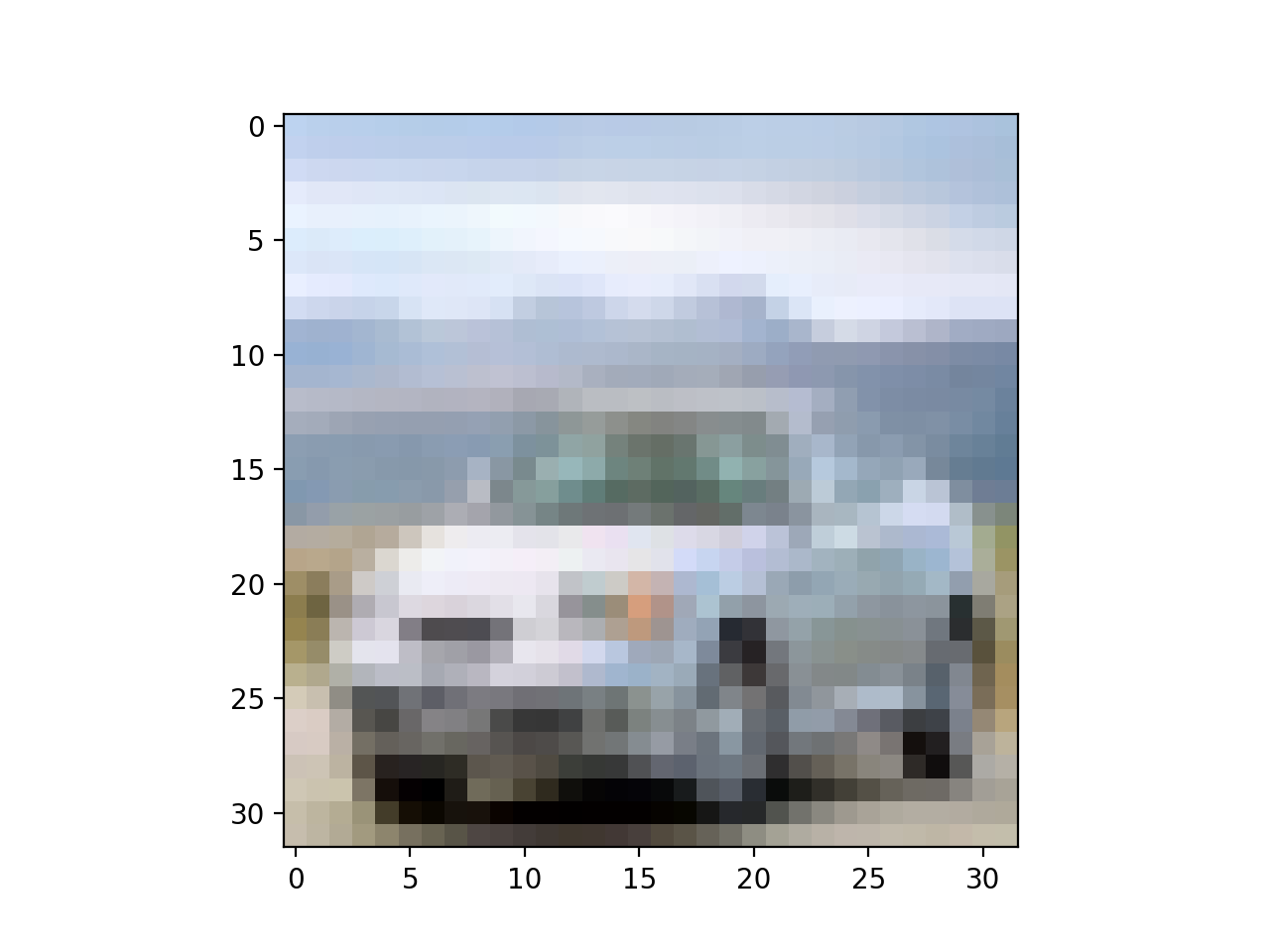}
  \vskip -0.in
  \caption{LR: [0.13, 0.97] \\ DR: [0.08, 0.64] \\ EN: [0.11, 1.00]}
\end{subfigure}
\hfill
\begin{subfigure}{0.3\textwidth}
  \centering
  \includegraphics[width=\linewidth]{CIFARtest-2.png}
  \vskip -0.in
  \caption{LR: [0.82, 0.99] \\ DR: [0.84, 0.99] \\ EN: [0.99, 1.00]}
  \label{fig:1}
\end{subfigure}
\hfill
\begin{subfigure}{0.3\textwidth}
  \centering
  \includegraphics[width=\linewidth]{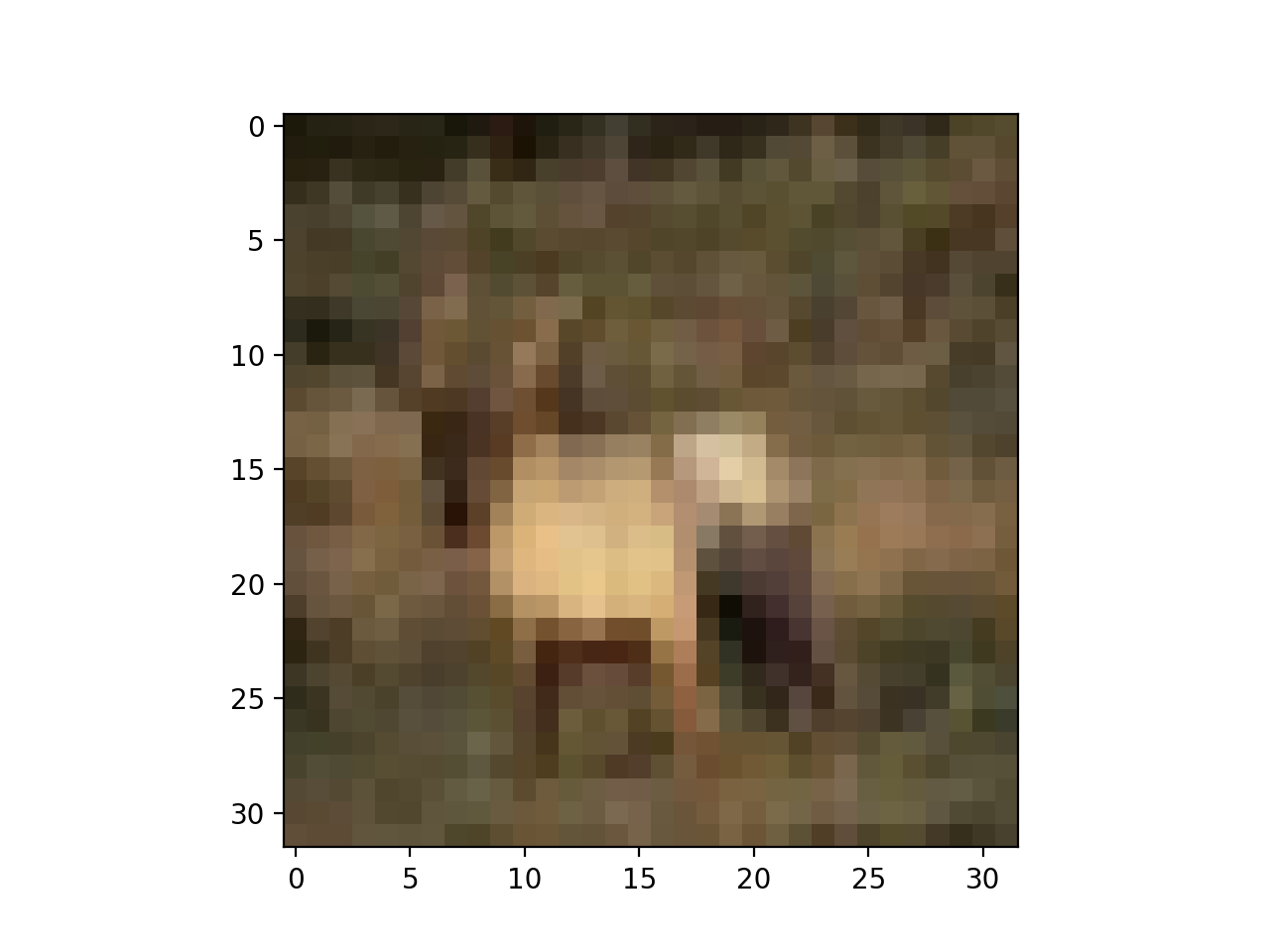}
  \vskip -0.in
  \caption{LR: [0.00, 0.69] \\ DR: [0.11, 0.40] \\ EN: [0.02, 0.43]]}
\end{subfigure} 
\caption{This figures gives 95$\%$ CIs for various training points (first column), test points (second column), and OoD points (third column). Each subcaption provides the CIs for the probability of class 1 (cars) made with DeepLR (LR), MC-dropout (DR), and ensembling (EN). All methods demonstrate greater uncertainty than in the MNIST example, which is sensible as the CIFAR data set is significantly harder. Our method is very uncertain for all OoD points, which is not always the case for MC-dropout (c and i) and ensembling (i). We also observe relatively uncertain predictions by all methods for some in-distribution points, notably (e) and (g).}
\label{fig: cifar10}
\end{figure}

Figure \ref{fig: cifar10} presents 95$\%$ confidence intervals for several training, test, and OoD points. Our method is uncertain for out-of-distribution inputs. However, contrary to the MNIST example, we also see that the model is uncertain for various in-distribution points. Figures \ref{fig: cifar10}(e) and \ref{fig: cifar10}(g) provide examples of such uncertain predictions. The exact reason why the model is uncertain for those inputs remains speculation. Possible explanations might be the open hood of the car in (e) or the large amount of blue sky in (g). An interesting avenue for future work would be to investigate what specific features cause DeepLR to output greater uncertainty.

\subsection{Adversarial example}
In addition to the previous experiments, we briefly tested how the method deals with adversarial examples \citep{goodfellow2014explaining}. Adversarial examples are modified inputs that are specifically designed to mislead the model, typically by adding small perturbations to the input. While virtually imperceptible to humans, these perturbations can dramatically alter the model's prediction.

We constructed adversarial versions of the two most confident test-inputs in Figure \ref{fig: cifar10} using the FGSM method \citep{goodfellow2014explaining}, which works by using the gradients of the networks' loss function with respect to the input to create an input that maximizes the loss. 

\begin{figure}[h!]
\centering
  \begin{minipage}{0.49\textwidth}
    \centering
   \textbf{Normal}
  \end{minipage}
    \begin{minipage}{0.49\textwidth}
    \centering
   \textbf{Adversarial}
  \end{minipage}
\begin{subfigure}{0.49\textwidth}
  \centering
  \includegraphics[width=0.8\linewidth]{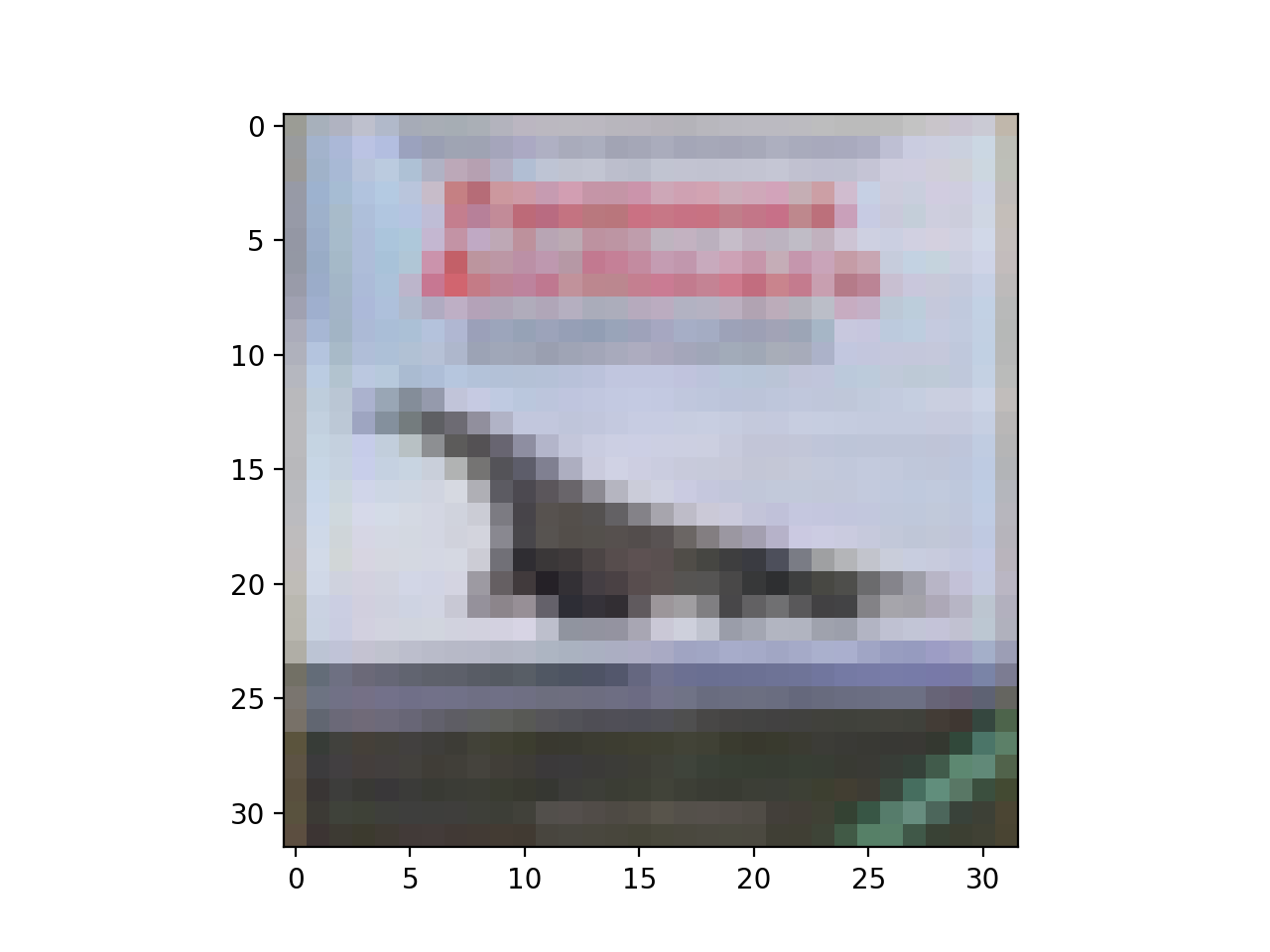}
  \caption{CI=[0.01, 0.14]}
  \label{fig:1}
\end{subfigure}
\begin{subfigure}{0.49\textwidth}
  \centering
  \includegraphics[width=0.8\linewidth]{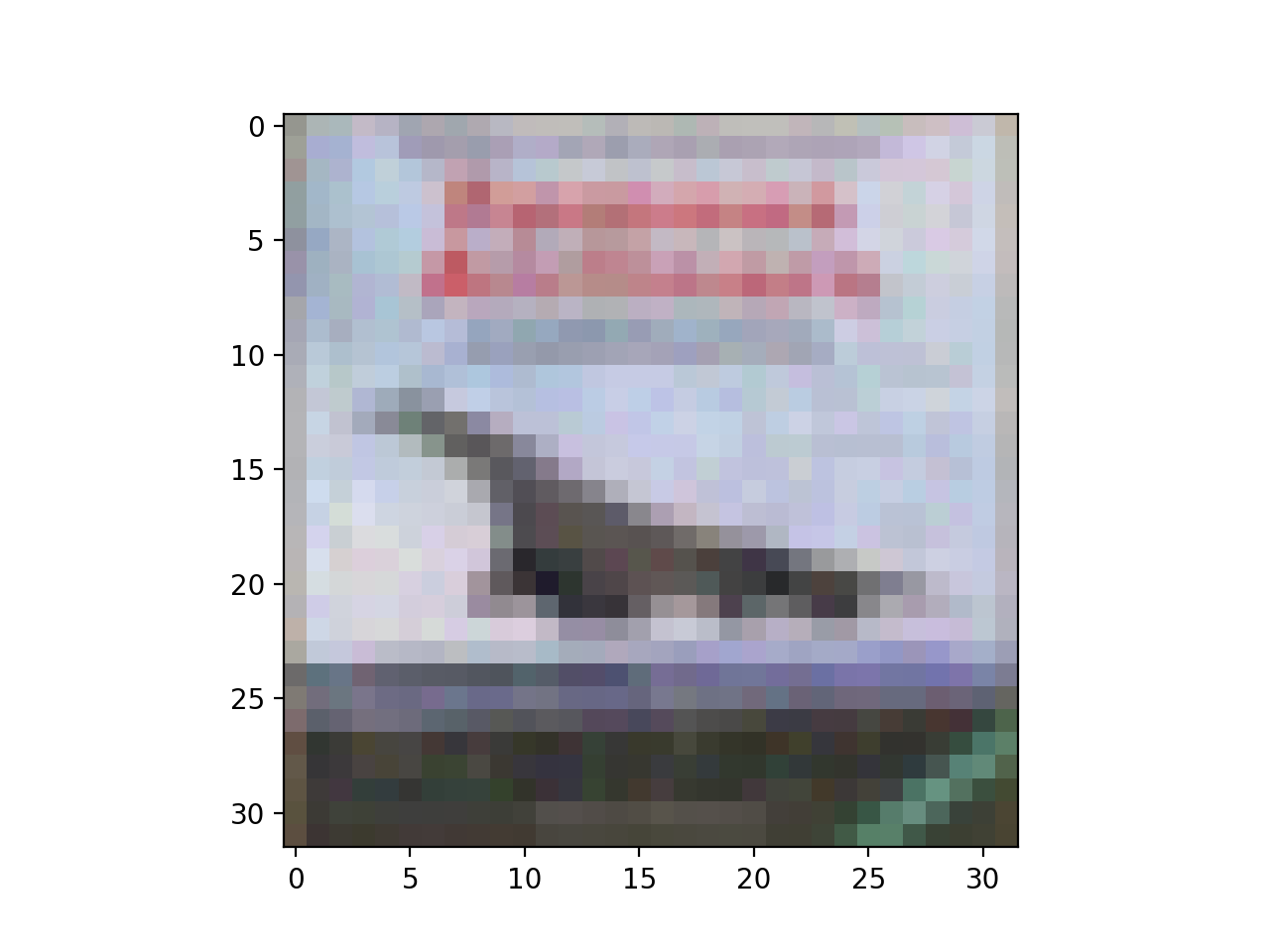}
  \caption{CI=[0.57, 1.00]}
  \label{fig:1}
\end{subfigure}
\vskip -0.1in
\begin{subfigure}{0.49\textwidth}
  \centering
  \includegraphics[width=0.8\linewidth]{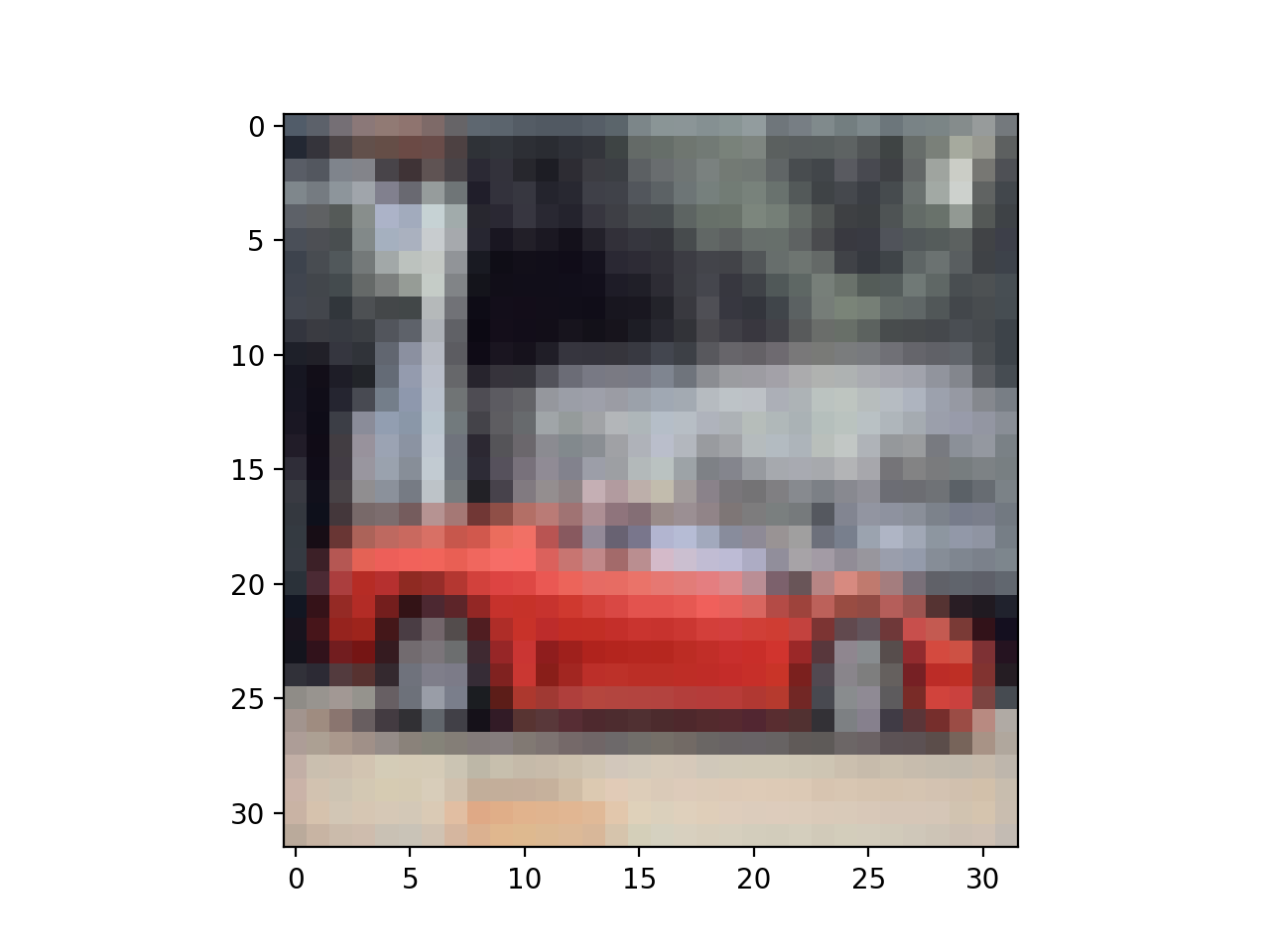}
  \caption{CI=[0.82, 0.99]}
  \label{fig:1}
\end{subfigure}
\begin{subfigure}{0.49\textwidth}
  \centering
  \includegraphics[width=0.8\linewidth]{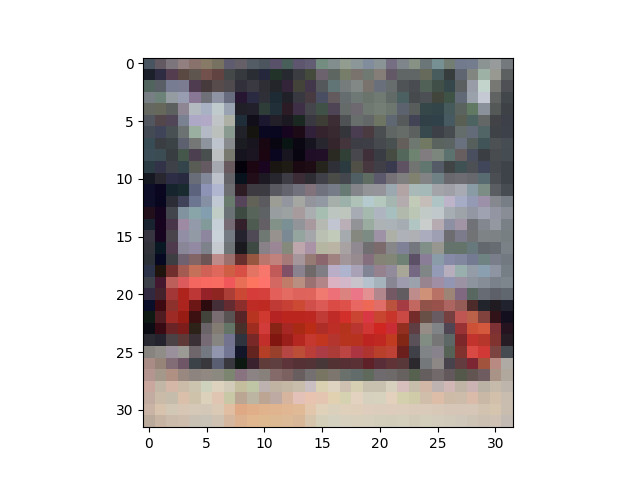}
  \caption{CI=[0.00, 0.77]}
  \label{fig:1}
\end{subfigure}
\caption{Adversarial inputs generated by the FGSM method result in more uncertain predictions. The same network was used as for the CIFAR example illustrated in Figure \ref{fig: cifar10}. The intuition behind the larger CIs is as follows. While very similar to humans, the adversarial examples are significantly different to a neural network. This allows the network to change the prediction for the adversarial example without changing the predictions of the training data, resulting in larger confidence intervals.}
\label{fig: cifaradv}
\end{figure}

As illustrated in Figure \ref{fig: cifaradv}, DeepLR demonstrates a higher uncertainty for the two adversarial inputs. This effect can be explained as follows. Although very similar to a human observer, an adversarial example is significantly different to a neural network. The difference was so large that both adversarial examples were wrongly classified by the network. Since these adversarial examples differ significantly from the training data, the network can more easily change the prediction at this location without changing the other predictions and thus the likelihood of the training data too much. This results in much larger confidence intervals. 

These results provide an encouraging sign that our method could also be able to handle adversarial examples, offering the unique capability to create confidence intervals, detect OoD examples, and offer robustness against adversarial examples, all with a single method. Unfortunately, the high computational cost of the method prevents more large-scale comparisons with other methods to more firmly establish this potential.

\FloatBarrier
\section{Discussion and conclusion} \label{discussionconclusion}
In this paper, we demonstrated the potential of a likelihood-ratio-based uncertainty estimate for  neural networks. This approach is capable of producing asymmetric confidence intervals that are better motivated in cases where we have fewer data points than parameters, i.e. most deep learning applications. 

The experimental results verify the theoretical advantages of a likelihood-ratio-based approach. The intervals are larger in regions with fewer data points, can get asymmetric in biased regions, and get larger for OoD and adversarial outputs. However, not being specifically designed for OoD detection or robustness against adversarial attacks, we do not claim it to be competitive in this regard against tailor-made alternatives.

While we made an effort to reduce it, our method still has some variance and can produce slightly different intervals upon repetition due to the randomness of the optimization procedure. This effect is greater for larger data sets where small differences can have a large effect on the test statistic.

Furthermore, it is essential that the model is well specified. This is the case for any model and not specific to our method. If the true density $p_{\theta_{0}}$ cannot be reached, the resulting confidence intervals will surely be wrong. A model can be miss specified if it is overly regularized or if incorrect distributional assumptions are made (e.g. incorrectly assuming Gaussian noise). 

In its current form, the high computational cost makes DeepLR unsuitable for many deep learning applications. A self-driving car that is approaching a cross-section cannot stop and wait for an hour until it has an uncertainty estimate. Nevertheless, a trustworthy uncertainty estimate may be critical in certain situations, or only a limited number of confidence intervals may be required. For instance, for medical applications, the extra computational time may be worthwhile. Alternatively, for some applications within astrophysics, only very few confidence intervals may be needed. If only a single interval is needed, for instance, our method is cheaper than an ensemble.

\subsection{Future work}
Overall, our findings highlight the potential of a likelihood-ratio-based approach as a new branch of uncertainty estimation methods. We hope that our work will inspire further research in this direction. Several areas for improvement include:
\begin{itemize}
	\item Reduced computational cost: A clear limitation of the current implementation is the cost. For every confidence interval, we need to train two additional networks. We acknowledge that this is infeasible for many - although not all - applications and we hope that the proof of concept in this paper motivates further research in this direction that may result in reduced computational cost.
	\item Improved approximation of the test statistic: The calculation of the test statistic uses an approximation for the second term in equation \eqref{eq: teststatistic}. Further research could focus on finding better and possibly cheaper approximations. It may also be worthwhile to investigate the use of a Bartlett correction.
	\item Application to other machine-learning models: We applied this approach to neural networks. However, the methodology should also be applicable to other models, for example random forests. Especially for models that are relatively cheap to train, this approach could be very promising.
	\item Development of the theory on the distribution of the test statistic: It would be interesting to develop the theory surrounding the distribution of the test statistic in greater generality, possibly also when explicitly considering a regularization term. We constructed a reparametrization that showed that, under some assumptions, the test statistic has a  $\chi^{2}(1)$ distribution. It would be interesting to study these assumptions further.
	\item A better understanding of what causes DeepLR to become uncertain: We saw, for example, that various planes and cars had rather large accompanying confidence intervals. It would be interesting to study what causes certain input to become more uncertain than others.
\end{itemize}

\FloatBarrier
\bibliographystyle{apalike}
\bibliography{../../references3}


\newpage
\appendix 
\section{Proof of theorem} \label{prooftestthm}
\setcounter{theorem}{0}
\begin{theorem} \label{thm: testdisttheorem}
Assume that $\Theta$ contains an open subset around $\theta_0$ in $\R^p$. Let $l(\mathcal{D};\theta)$ be the loglikelihood of the data given $\theta$. Define $\Theta_{0}(c) = \{ \theta \in \Theta \mid f_{\theta}(X_{0}) = c \}$ and denote the tangent space of $\Theta_{0}(c)$ at $\theta_{0}$ with $T_{\theta_{0}}\Theta_{0}(c)$. Also define $\hat{\theta}$ as the Maximum Likelihood Estimator (MLE) of $\theta_0$ and $\hat{\theta}_0$ as the MLE when we restrict our parameterspace to $\Theta_{0}(c)$.

We assume:
\begin{itemize}
\item[A1:] $l(\mathcal{D};\theta)$ is three times continuously differentiable in a neighbourhood of $\theta_0$,
\item[A2:] $\hat{\theta}$ = $\theta_{0} + o_p(n^{-1/4})$ and $\hat{\theta}_0$ = $\theta_{0} + o_p(n^{-1/4})$,
\item[A3:] There exists a vector $\hat{n}\in \mathbb{R}^{p}$ such that $h_{0}^{T}\ddot{l}(\theta_{0})\hat{n} = O_p(n\|h_0\|^2)$ for all $h_{0} \in (\Theta_{0}(c) - \theta_0)$. Also, $\n$ is transversal to $T_{\theta_0}\Theta_{0}(c)$.
\item[A4:] There exists a constant $i>0$ such that $\frac{1}{n}\n^{T}\ddot{l}(\theta_{0})\hat{n} = -i + o_p(1)$.
\end{itemize}
Under these assumptions, the test statistic
\[
T(c) = 2\bigg(\sup_{\Theta}l(\mathcal{D};\theta)) - \sup_{\Theta_{0}(c)}l(\mathcal{D};\theta)\bigg)
\]
converges in distribution to a $\chi^{2}(1)$ distribution.
\end{theorem}
We will first prove this theorem and then comment on the assumptions we make.
\proof{
 Our strategy is to construct a reparametrization
\[
\Psi: \Theta_{0}(c) \times \mathbb{R} \to \Theta : (\tilde{\theta}, t) \mapsto \Psi(\tilde{\theta}, t), \quad \text{such that}
\]
\[
L(\mathcal{D}; \Psi(\tilde{\theta}, t)) \approx L(\mathcal{D};\tilde{\theta}) \phi(\mathcal{D};t),
\]
with high probability for $\tilde{\theta}\in \Theta_{0}$ close to $\theta_{0}$, for some function $\phi$. 
In other words, we use a parametrization such that the likelihood factorizes in a part that depends on $\tilde{\theta}\in \Theta_{0}(c)$, and a part that depends on $t \in \mathbb{R}$. If we can construct such a parametrization, then $T(c)$ will reduce to 2 times the loglikelihood-ratio of a one-dimensional model, which is known to converge in distribution to a $\chi^{2}(1)$ distribution. 

We use the following parametrization:
\[
\theta = P(\theta) + t\n,
\]
with $P(\theta)$ being the projection onto $\Theta_0$. In a neighbourhood of $\theta_0$ this projection (i.e., the closest point in $\Theta_0$ to $\theta$) is uniquely defined and since $\n$ is not tangent to $T_{\theta_{0}}\Theta_{0}(c)$, $t$ is also unique. We define $\hat{h}_0$ and $\hat{h}_0^{(0)}$ such that
\[ \hat{h}_0 = P(\hat{\theta})-\theta_0\ \ \mbox{and}\ \ \hat{h}_0^{(0)}=\hat{\theta}_0 - \theta_{0}.\]
See Figure \ref{fig: proofvisualization} for a visualization of the notation. Condition A2 implies that $ \|\hat{h}_0\|=o_p(n^{-1/4})$ and $\|\hat{h}_0^{(0)}\|=o_p(n^{-1/4})$. We also define $\hat{t}$ such that $\hat{\theta} = P(\hat{\theta}) + \hat{t}\n$. Finally, we define $\bar{t}$ by
\[ \bar{t} = \argmax_{t\in \R}\left( \dot{l}(\theta_{0})t\n + \frac{1}{2}t^{2}\n^{T}\ddot{l}(\theta_{0})\n \right)= -\frac{\dot{l}(\theta_{0})\n}{\n^{T}\ddot{l}(\theta_{0})\n}.\]
\begin{figure}[h]
\centering
\includegraphics{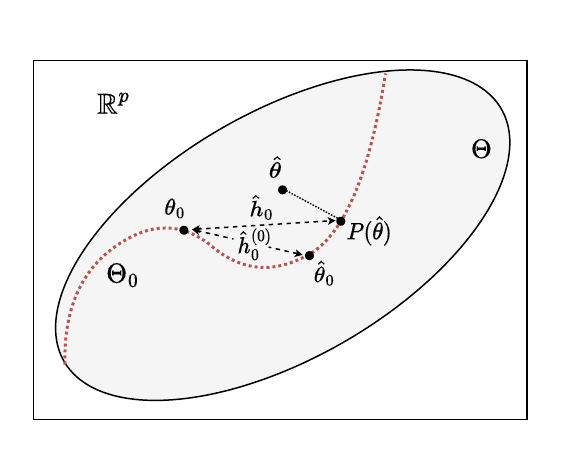}
\caption{Visualization of the notation used in the proof. The shaded area represents $\Theta$ and the dotted red line $\Theta_{0}(c)$.}	
\label{fig: proofvisualization}	
\end{figure}

By condition A4, this is indeed a maximum for large enough $n$. The loglikelihood is a sum of $n$ i.i.d. random variables, and the expectation has its maximum at $\theta_0$. Therefore, $l(\theta_0)$ and its derivatives are all $O_p(n)$, but since $\dot{l}(\theta_0)$ is a sum of i.i.d. random variables with expectation $0$, we see that $\dot{l}(\theta_0)=O_p(n^{1/2})$; this implies that $\bar{t}=O_p(n^{-1/2})$. In what follows, we drop the dependence of the loglikelihood on the data for notational simplicity. A Taylor expansion of the loglikelihood and using the fact that $\hat{\theta}$ and $\hat{\theta}_0$ are MLE's yield the following inequalities:
\begin{align}
\nonumber l(\hat{\theta}) &= l(\theta_{0} + \hat{h}_{0} + \hat{t}\n)\\ 
\nonumber & \geq l(\theta_0 + \hat{h}^{(0)}_0 + \bar{t}\n)\\
\nonumber &=l(\theta_{0} + \hat{h}^{(0)}_0) + \dot{l}(\theta_{0} + \hat{h}^{(0)}_0)\bar{t}\n+ \frac{1}{2}\bar{t}^{2}\n^{T}\ddot{l}(\theta_{0} + \hat{h}^{(0)}_0)\n + O_p(n\bar{t}^{3})\\
\nonumber &= l(\theta_{0} + \hat{h}^{(0)}_0) + \dot{l}(\theta_{0})\bar{t}\n + {\hat{h}^{(0)\,T}_0}\ddot{l}(\theta_{0})\bar{t}\n + \frac{1}{2}\bar{t}^{2}\n^{T}\ddot{l}(\theta_{0})\n  \\
\label{ineq:1.1}& \quad \quad \quad \quad \quad \quad \quad \quad \quad \quad\quad + O_p(n\bar{t}^{3} + n\bar{t}^{2}||\hat{h}^{(0)}_0|| + n\bar{t}||\hat{h}^{(0)}_0||^{2}),
\end{align}
and
\begin{align}
\nonumber l(\hat{\theta}) &= l(\theta_{0} + \hat{h}_{0} + \hat{t}\n)\\
\nonumber &=l(\theta_{0} +  \hat{h}_{0}) + \dot{l}(\theta_{0} + \hat{h}_{0})\hat{t}\n+ \frac{1}{2}\hat{t}^{2}\n^{T}\ddot{l}(\theta_{0} +  \hat{h}_{0})\n + O_p(n\hat{t}^{3})\\
\nonumber &\leq  l(\theta_{0} + \hat{h}^{(0)}_0) + \dot{l}(\theta_{0})\hat{t}\n +  \hat{h}_{0}^T\ddot{l}(\theta_{0})\hat{t}\n + \frac{1}{2}\hat{t}^{2}\n^{T}\ddot{l}(\theta_{0})\n \\
\label{ineq:2.1}& \quad \quad \quad \quad \quad \quad \quad \quad \quad \quad \quad \quad \quad \quad \quad \quad + O_p(n\hat{t}^{3} + n\hat{t}^{2}||\hat{h}_{0}|| + n\hat{t}||\hat{h}_{0}||^{2})\\
\nonumber &\leq  l(\theta_{0} + \hat{h}^{(0)}_0) + \dot{l}(\theta_{0})\bar{t}\n +  \hat{h}_{0}^T\ddot{l}(\theta_{0})\hat{t}\n + \frac{1}{2}\bar{t}^{2}\n^{T}\ddot{l}(\theta_{0})\n \\
\label{ineq:2.2}& \quad \quad \quad \quad \quad \quad \quad \quad \quad \quad \quad \quad \quad \quad \quad \quad + O_p(n\hat{t}^{3} + n\hat{t}^{2}||\hat{h}_{0}|| + n\hat{t}||\hat{h}_{0}||^{2}).
\end{align}

Combining \eqref{ineq:2.1} and \eqref{ineq:1.1}, we see that $\hat{t}=O_p(\bar{t})=O_p(n^{-1/2})$. Using A2, we also see that all terms in the remainder are $o_p(1)$. Because of A3, this is also true for ${\hat{h}^{(0)\,T}_0}\ddot{l}(\theta_{0})\bar{t}\n$ and $\hat{h}_{0}^T\ddot{l}(\theta_{0})\hat{t}\n$. Using \eqref{ineq:1.1} and \eqref{ineq:2.2}, our test statistic therefore reduces to
\begin{align*}
	T(c) &= 2\bigg(\sup_{\theta \in \Theta}l(\theta) - \sup_{\theta \in \Theta_{0}(c)}l(\theta)\bigg) \\
	& = 2(l(\hat{\theta}) - l(\hat{\theta}_0)) \\
	& = 2\dot{l}(\theta_{0})\bar{t}\n +  \bar{t}^{2}\n^{T}\ddot{l}(\theta_{0})\n +o_p(1)\\
	& = \frac{(\dot{l}(\theta_{0})\n)^2}{n\cdot i} + o_p(1).
\end{align*}

By the central limit theorem and a well-known property of the loglikelihood relating the variance of the derivative to the expectation of the second derivative, $\frac{1}{\sqrt{n}}\frac{d}{dt}l(\theta_{0}+t\n)\rvert_{t=0}$ converges in distribution to a normal distribution with mean zero and covariance $i\in \mathbb{R}$. We conclude that $T(c)$ weakly converges to a $\chi^{2}(1)$ distribution. $\square$
}
\newline 
\\
Assumption A1 is necessary for the Taylor expansion, and A2 is important for localization, but notice that the rate of convergence in directions other than $\n$ can be slower than parametric. A4 mainly implies that the Fisher information of the relevant one-dimensional model is not $0$. The more technical A3 requires more explanation: a natural choice would be to consider the expected $\ddot{l}(\theta_0)$ and pick $\n$ such that $\ddot{l}(\theta_0)\n$ is perpendicular to $\Theta_{0}(c)$ at $\theta_0$. This does not require that the full second derivative matrix is invertible (a strong requirement when the parameter space is very high dimensional), since it only needs to solve one vector equation. The intuition is that the second derivative only needs to behave well in a direction transversal to $\Theta_{0}(c)$.\\

We acknowledge that we cannot prove that these assumptions hold in practice. It could happen, for instance, that $T_{\theta_{0}}\Theta_{0}(c) = \Theta$. In this case, we are overparameterized to such an extent that the restriction $f_{\theta}(X_{0})=c$ has no effect on the likelihood, leading to a test statistic of 0, which clearly is not $\chi^{2}(1)$ distributed. Our resulting confidence interval, however, is still conservative in this case since we will not reject the null hypothesis when the test statistic is zero. 
\end{document}